\documentclass[10pt,journal,compsoc]{IEEEtran}
\usepackage{graphicx}
\usepackage{balance}  % for  \balance command ON LAST PAGE  (only there!)

\usepackage{pslatex,epsfig}
\usepackage{multirow}
\usepackage{upquote}
\usepackage{url}
\usepackage{xcolor}
\usepackage{epstopdf}
\usepackage{amssymb}
\usepackage{amsmath}
\usepackage{enumerate}
\usepackage{subfigure}
\usepackage{graphicx}
\usepackage{multirow}
\usepackage[T1]{fontenc}
\usepackage[linesnumbered,ruled]{algorithm2e}
\usepackage{euler}
\usepackage{color}

\newcommand{\eat}[1]{}

\begin{document}

\title{The Gap of Semantic Parsing: A Survey on Automatic Math Word Problem Solvers}

\author{Dongxiang~Zhang, Lei~Wang, Luming~Zhang, Bing~Tian~Dai and Heng~Tao~Shen 
%~\IEEEmembership{Fellow,~IEEE} %$\dag$
\IEEEcompsocitemizethanks{
\IEEEcompsocthanksitem D. Zhang and L. Zhang  are with the College of Computer Science and Technology, Zhejiang University, China. Emails: \{zhangdongxiang37,zglumg\}@gmail.com; 

\IEEEcompsocthanksitem L. Wang and H. T. Shen  are with the Center for Future Media and School of Computer Science and Engineering, University of Electronic Science and Technology of China. Emails:  demolei@outlook.com; shenhengtao@hotmail.com

\IEEEcompsocthanksitem B. T. Dai is with School of Information Systems, Singapore Management University.  Email:btdai@smu.edu.sg
\IEEEcompsocthanksitem Corresponding Author:  Heng Tao Shen.
}
}

\IEEEtitleabstractindextext{%
\begin{abstract}
Solving mathematical word problems (MWPs) automatically is challenging, primarily due to the semantic gap between human-readable words and machine-understandable logics. Despite the long history dated back to the $1960$s, MWPs have regained intensive attention in the past few years with the advancement of Artificial Intelligence (AI). Solving MWPs successfully is considered as a milestone towards general AI. Many systems have claimed promising results in self-crafted and small-scale datasets. However, when applied on large and diverse datasets, none of the proposed methods in the literature achieves high precision, revealing that current MWP solvers still have much room for improvement. This motivated us to present a comprehensive survey to deliver a clear and complete picture of automatic math problem solvers. In this survey, we emphasize on algebraic word problems, summarize their extracted features and  proposed techniques to bridge the semantic gap, and compare their performance in the publicly accessible datasets. We  also cover automatic solvers for other types of math problems such as geometric problems that require the understanding of diagrams. Finally, we identify several emerging research directions for the readers with interests in MWPs.
\end{abstract}

% Note that keywords are not normally used for peerreview papers.
}

\maketitle

\eat{

}
\IEEEdisplaynontitleabstractindextext

\IEEEpeerreviewmaketitle

%====================================================================================

%================================================================================
\section{Introduction}
\label{sec:intro}
Designing an automatic solver for mathematical word problems (MWPs) has a long history dated back to the $1960$s~\cite{bobrow,Feigenbaum:1963:CT:601134,Charniak:1969:CSC:1624562.1624593}, and continues to attract intensive research attention. In the past three years, more than $40$ publications on this topic have emerged in the premier venues of artificial intelligence. The problem is particularly challenging because there remains a wide semantic gap to parse the human-readable words into machine-understandable logics so as to facilitate quantitative reasoning. Hence, MWPs solvers are broadly considered as good test beds to evaluate the intelligence level of agents in terms of natural language understanding~\cite{DBLP:conf/aaai/Clark15,DBLP:journals/aim/ClarkE16} and the successful solving of MWPs would constitute a milestone towards general AI.

\begin{figure}[t]
\centering
\includegraphics[width=75mm, keepaspectratio]{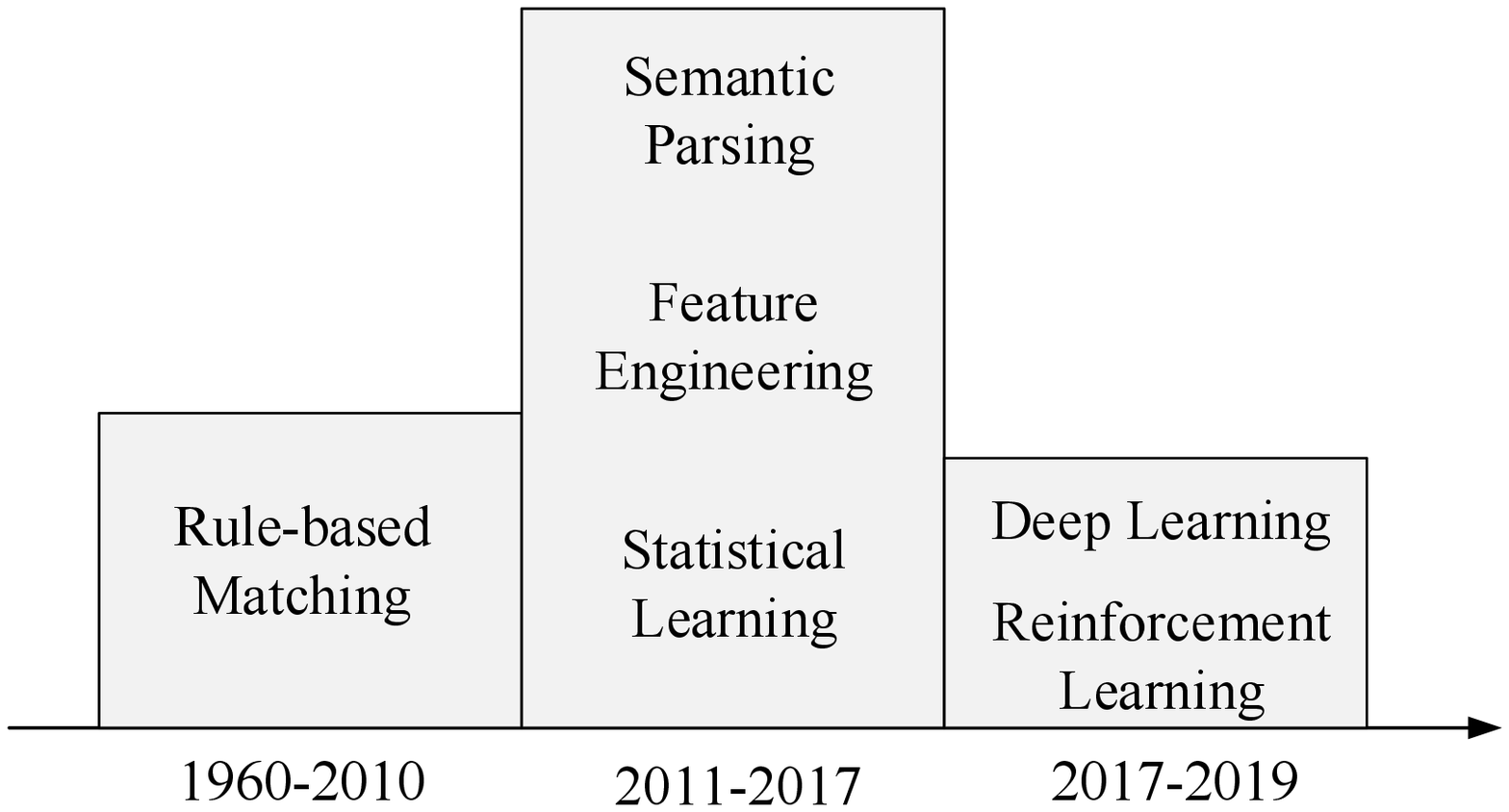} 
\caption{Technology evolving trend in solving MWPs.}
\label{fig:tech-trend}
\end{figure}

We categorize the evolution of MWP solvers into three major stages according to the technologies behind them, as depicted in Figure~\ref{fig:tech-trend}. In the first pioneering stage, roughly from the year $1960$ to $2010$, systems such as STUDENT~\cite{bobrow}, DEDUCOM~\cite{Slagle:1965:EDQ:365691.365960}, WORDPRO~\cite{Fletcher1985} and ROBUST~\cite{2007math1393B}, manually craft rules and schemas for pattern matchings. Thereupon, these solvers heavily rely on human interventions and can only resolve a limited number of scenarios that are defined in advance. Those early efforts for automatic understanding of natural language mathematical problems have been thoroughly reviewed in~\cite{Mukherjee:2008:RMA:1612882.1612895}. We exclude them from the scope of this survey paper and focus on the recent technology developments that have not been covered in the previous survey~\cite{Mukherjee:2008:RMA:1612882.1612895}.

%and refer interested readers to~\cite{Mukherjee:2008:RMA:1612882.1612895} for a thorough review on the early efforts for automatic understanding of natural language mathematical problems. 

In the second stage, MWP solvers made  use of semantic parsing~\cite{DBLP:conf/ijcai/GoldwasserR11,DBLP:conf/emnlp/KwiatkowskiCAZ13}, with the objective of mapping the sentences from problem statements into structured logic representations so as to facilitate quantitative reasoning. It has regained considerable interests from the academic community, and a booming number of methods have been proposed in the past years. These methods leveraged various strategies of feature engineering and statistical learning for performance boosting. The authors of these methods also claimed promising results in their public or manually harvested datasets. In this paper, one of our tasks is to present a comprehensive review on the proposed methods in this stage. The methods will be initially organized according to the sub-tasks of MWPs which they were designed to solve, such as arithmetic word problem (in Section~\ref{sec:arithmetic}), equation set word problem (in Section~\ref{sec:eqn-set}) and geometric word problem (in Section~\ref{sec:geometric}). We then examine the proposed techniques in each sub-task with a clear technical organization and accountable experimental evaluations.

MWP solvers in the third stage were originated from an empirical work~\cite{DBLP:conf/acl/HuangSLYM16}. Its experimental results on a large-scale and diversified dataset showed that the status of MWP solvers was not as optimistic as they claimed to be. In fact, the accuracies of many approaches dropped sharply and there is a great room for improvement in this research area. To design more accurate and robust solutions, the subsequent publications are forked into two directions. One is to continue refining the technology of semantic parsing. For instance, Huang et al. proposed a new type of semantic representation to conduct fine-grained inference~\cite{DBLP:conf/emnlp/HuangSYL17}. The other direction attempts to exploit the advantages of deep learning models, with the availability of large-scale training datasets. This is an emerging research direction for MWP solvers and we observed multiple instances, including Deep Neural Solver~\cite{DBLP:conf/emnlp/WangLS17}, Seq2SeqET~\cite{DBLP:conf/emnlp/WangWCZL18}, StackDecoder~\cite{DBLP:journals/corr/abs-1811-00720}, MathDQN~\cite{DBLP:conf/aaai/Wang18}, CASS~\cite{DBLP:conf/coling/HuangLLY18}, T-RNN~\cite{DBLP:conf/aaai/Wang19}. These models represent a new technology trend in the research topic of MWP solvers and will be paid special attention in the survey.
%Various forms of deep neural networks such as recurrent neural networks, deep reinforcement learning and recursive
% using recurrent neural network to generate the equation template, and  relying on the deep reinforcement learning framework. It is expected to witness more and more deep learning based methods to solve MWPs in the future.

To sum up, we present a comprehensive survey to review the MWP solvers proposed in recent years. Researchers in the community can benefit from this survey in the following ways:
\begin{enumerate}
\item We provide a wide coverage on the math word problems, including arithmetic word problem, equation set problem, geometry word problem and miscellaneous sub-tasks related to automatic math solvers. The practitioners can easily identify all relevant approaches for performance evaluations. We observed that the unawareness of relevant competitors occurs occasionally in the past literature. We are positive that the availability of our survey can help avoid such unawareness.

\item The solvers designed for arithmetic word problems (AWP) with only one unknown variable and equation set problems (ESP) with multiple unknown variables are often not differentiated by previous works. In fact, the methods proposed for ESP are more general and can be used to solve AWP. In this survey, we clearly identify the difference and organize them in separate sections. %We also generalize their methods in terms of the number of operands and the type of operators they are capable to support.

\item Feature engineering plays a vital role to bridge the gap of semantic parsing. Almost all MWP solvers state their strategies of crafting effective features, resulting in a very diversified group of features, and there is a lack of clear organization among these features. In this survey, we will be the first to summarize all these proposed features in Table~\ref{tbl:common-feature}.
% so that readers can conveniently find out the popular ones.

\item  As for the fairness on performance evaluations, ideally, there should be a benchmark dataset well accepted and widely adopted by the MWP research community, just like ImageNet~\cite{imagenet_cvpr09} for visual object recognition and VQA~\cite{7410636,DBLP:conf/cvpr/GoyalKSBP17} for visual question answering. Unfortunately, we observed that many approaches tend to compile their own datasets to verify their superiorities, resulting in missing of relevant competitors as mentioned above. Tables~\ref{tbl:exp-arithmetic} and~\ref{tbl:dataset-eqn-set} integrate the results of the existing methods on all public datasets. After collecting the accuracies that have been reported in the past literature, we observed many empty cells in the table. Each empty cell refers to a missing experiment on a particular algorithm and dataset. In this survey, we make our best efforts to fill the missing results by conducting a number of additional experiments, that allow us to provide a more comprehensive comparison and explicit analysis.
\end{enumerate}

The remainder of the paper is organized as follows. We first review the arithmetic word problem solvers in Section~\ref{sec:arithmetic}, followed by equation set problem solvers in Section~\ref{sec:eqn-set}. Since feature engineering deserves special attention, we summarize the extracted features as well as the associated pre-processing techniques in Section~\ref{sec:feature}. The geometric word problem solvers are reviewed in Section~\ref{sec:geometric}.  We also cover  miscellaneous automatic  solvers related to math problems in Section~\ref{sec:misc}. We conclude the paper and point out several future directions in MWPs that are worth examination in the final section.

\section{Arithmetic Word Problem Solver}\label{sec:arithmetic}
The arithmetic word problems are targeted for elementary school students. The input is the text description for the math problem, represented in the form of a sequence of $k$ words $\langle w_0, w_1,\ldots,w_k\rangle$. There are $n$ quantities $q_1,q_2,\ldots,q_n$ mentioned in the text and an unknown variable $x$ whose value is to be resolved. Our goal is to extract the relevant quantities and map this problem into an arithmetic expression $E$ whose evaluation value provides the solution to the problem. There are only four types of fundamental operators $\mathcal{O}=\{+,-,\times,\div \}$ involved in the expression $E$.

\begin{figure}[h]
\centering
\includegraphics[width=68mm, keepaspectratio]{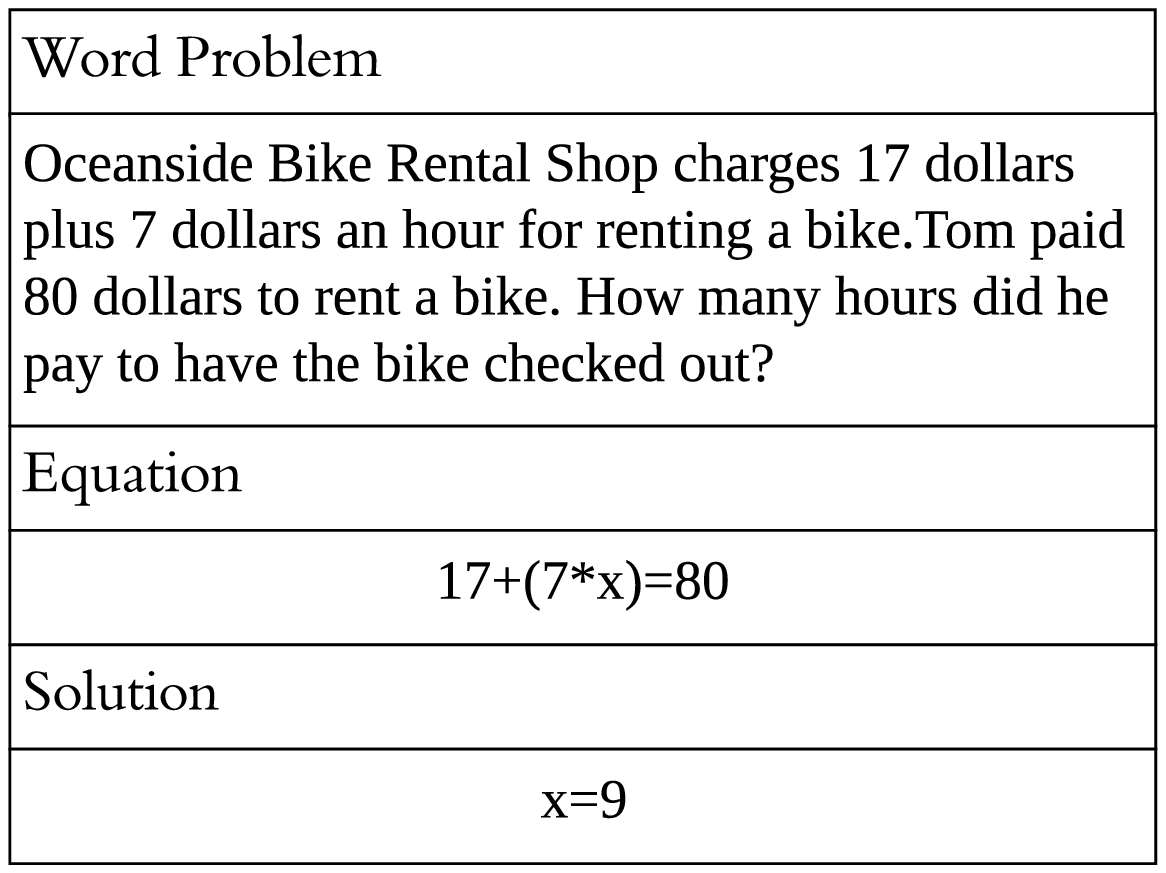} 
\caption{An example of arithmetic word problem.}
\label{fig:example-arithmetic}
\end{figure}

An example of arithmetic word problem is illustrated in Figure~\ref{fig:example-arithmetic}. The relevant quantities to be extracted from the text include $17$, $7$ and $80$. The number of hours spent on the bike is the unknown variable $x$. To solve the problem, we need to identify the correct operators between the quantities and their operation order such that we can obtain the final equation $17+7x=80$ or expression $x=(80-17)\div 7$ and return $9$ as the solution to this problem.

In this section, we consider feature extraction as a black box and focus on the high-level algorithms and models. The details of feature extraction will be comprehensively present in Section~\ref{sec:feature}. We classify existing algebra word problem solvers into three categories: rule-based, statistic-based, tree-based and deep learning (DL)-based methods.

\subsection{Rule-based Methods}
The early approaches to math word problems are  rule-based systems based on hand engineering. Published in $1985$, WORDPRO~\cite{Fletcher1985} solves one-step arithmetic problems. It predefines four types of schemas, including \textit{change-in}, \textit{change-out}, \textit{combine} and \textit{compare}. The problem text is transformed into a set of propositions and the answer is derived with simple reasoning based on the propositions. Another system ROBUST, developed by Bakman~\cite{2007math1393B}, could understand free-format multi-step arithmetic word problems. It further expands the \textit{change} schema of WORDPRO~\cite{Fletcher1985} into six distinct categories. The problem text is split into sentences and each sentence is mapped to a proposition. Yun et al. also proposed to use schema for multi-step math problem solving~\cite{Yun2010}. However, the implementation details are not explicitly revealed. Since these systems have been out of date, we only provide such a brief overview to cover the representative ones. Readers can refer to~\cite{Mukherjee:2008:RMA:1612882.1612895} for a comprehensive survey of early rule-driven systems for automatic understanding of natural language math problems.

\subsection{Statistic-based Methods}
The statistic-based methods leverage traditional machine learning models to identify the entities, quantities and operators from the problem text and yield the numeric answer with simple logic inference procedure. The scheme of quantity entailment proposed in~\cite{DBLP:journals/tacl/RoyVR15} can be used to solve arithmetic problems with only one operator. It involves three types of classifiers to detect different properties of the word problem. The \textit{quantity pair classifier} is trained to determine which pair of quantities would be used to derive the answer. The \textit{operator classifier} picks the operator $op\in\{+,-,\times,\div\}$ with the highest probability. The \textit{order classifier} is relevant only for problems involving subtraction or division because the order of operands matters for these two types of operators. With the inferred expression, it is straightforward to calculate the numeric answer for the simple math problem.

To solve math problems with multi-step arithmetic expression, the statistic-based methods require more advanced logic templates. This usually incurs additional overhead to annotate the text problems and associate them with the introduced template. As an early attempt, ARIS~\cite{DBLP:conf/emnlp/HosseiniHEK14} defines a logic template named \textit{state} that consists of a set of entities, their containers, attributes, quantities and relations. For example, ``\textit{Liz has $9$ black kittens}'' initializes the number of \textit{kitten} (referring to an entity) with \textit{black} color (referring to an attribute) and belonging to \textit{Liz} (referring to a container). The solution splits the problem text into fragments and tracks the update of the states by verb categorization. More specifically, the verbs are classified into seven categories: \textit{observation}, \textit{positive}, \textit{negative}, \textit{positive transfer}, \textit{negative transfer}, \textit{construct} and \textit{destroy}. To train such a classifier, we need to annotate each split fragment in the training dataset with the associated verb category. Another drawback of ARIS is that it only supports addition and subtraction. ~\cite{sundaram-khemani:2015:W15-59} follows a similar processing logic to ARIS. It predefines a corpus of logic representation named \textit{schema}, inspired by~\cite{2007math1393B}. The sentences in the text problem are examined sequentially until the sentence matches a schema, triggering an update operation to modify the number associated with the entities.

Mitra et al. proposed a new logic template named \textit{formula}~\cite{DBLP:conf/acl/MitraB16}. Three types of formulas are defined, including \textit{part whole}, \textit{change} and \textit{comparison}, to solve problems with addition and subtraction operators. For example, the text problem ``\textit{Dan grew $42$ turnips and $38$ cantelopes. Jessica grew $47$ turnips. How many turnips did they grow in total?}'' is annotated with the part-whole template: $\langle whole:x, parts:\{42,47\}\rangle$. To solve a math problem, the first step connects the assertions to the formulas. In the second step, the most probable formula is identified using the log-linear model with learned parameters and converted into an algebraic equation. 

Another type of annotation is introduced in~\cite{Liang16,DBLP:conf/ijcai/LiangHHLMS16} to facilitate solving a math problem. A group of \textit{logic forms} are predefined and the problem text is converted into the logic form representation by certain mapping rules. For instance, the sentence ``\textit{Fred picks $36$ limes}'' will be transformed into $verb(v_1,pick)$ $\&$ $nsubj(v_1,Fred)$ $\&$ $dobj(v_1,n_1)$ $\&$ $head(n_1,lime)$ $\&$ $nummod(n_1,36)$. Finally, logic inference is performed on the derived logic statements to obtain the answer. 

To sum up, these statistical-based methods have two drawbacks that limit their usability. First, it requires additional annotation overhead that prevents them from handling large-scale datasets. Second, these methods are essentially based on a set of pre-defined templates, which are brittle and rigid. It will take great efforts to extend the templates  to support other operators like multiplication and division. It is also not robust to diversified datasets. In the following, we will introduce the tree-based solutions, which are widely adopted and become the main-streaming solutions to arithmetic word problems.

\subsection{Tree-Based Methods}
\begin{figure}[h]
\centering
\includegraphics[width=63mm, keepaspectratio]{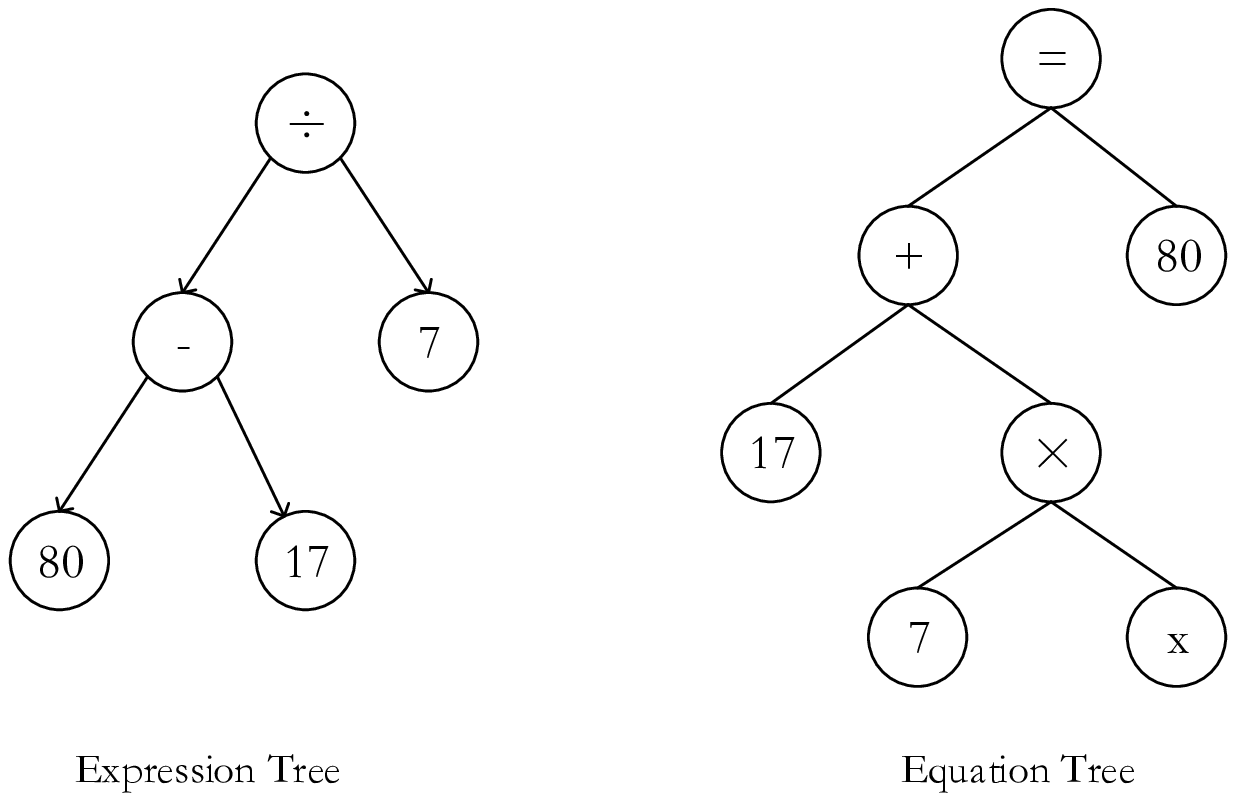} 
\caption{Examples of expression tree and equation tree for Figure~\ref{fig:example-arithmetic}.}
\label{fig:tree-examples}
\end{figure}

The arithmetic expression can be naturally represented as a binary tree structure such that the operators with higher priority are placed in the lower level and the root of the tree contains the operator with the lowest priority. The idea of tree-based approaches~\cite{DBLP:conf/emnlp/RoyR15,DBLP:journals/tacl/Koncel-Kedziorski15,DBLP:conf/aaai/RoyR17,DBLP:conf/aaai/Wang18} is to transform the derivation of the arithmetic expression to constructing an equivalent tree structure step by step in a bottom-up manner. One of the advantages is that there is no need for additional annotations such as equation template, tags or logic forms. Figure~\ref{fig:tree-examples} shows two tree examples derived from the math word problem in Figure~\ref{fig:example-arithmetic}. One is called \textit{expression tree} that is used in~\cite{DBLP:conf/emnlp/RoyR15,DBLP:conf/aaai/RoyR17,DBLP:conf/aaai/Wang18} and the other is called \textit{equation tree}~\cite{DBLP:journals/tacl/Koncel-Kedziorski15}. These two types of trees are essentially equivalent and result in the same solution, except that equation tree contains a node for the unknown variable $x$.

The overall algorithmic framework among the tree-based approaches consists of two processing stages. In the first stage, the quantities are extracted from the text and form the bottom level of the tree. The candidate trees that are syntactically valid, but with different structures and internal nodes, are enumerated. In the second stage, a scoring function is defined to pick the best matching candidate tree, which will be used to derive the final solution. A common strategy among these algorithms is to build a local classifier to determine the likelihood of an operator being selected as the internal node. Such local likelihood is taken into account in the global scoring function to determine the likelihood of the entire tree.

Roy et al.~\cite{DBLP:conf/emnlp/RoyR15} proposed the first algorithmic approach that leverages the concept of expression tree to solve arithmetic word problems. Its first strategy to reduce the search space is training a binary classifier to determine whether an extracted quantity is relevant or not.  Only the relevant ones are used for tree construction and placed in the bottom level. The irrelevant quantities are discarded. 
%After identifying the relevant quantities, a straightforward idea is to enumerate all the valid expression trees that use them as the leaf nodes. 
The tree construction procedure is mapped to a collection of simple prediction problems, each determining the lowest common ancestor operation between a pair of quantities mentioned in the problem. The global scoring function for an enumerated tree takes into account two terms. The first one, denoted by $\phi(q)$, is the likelihood of quantity $q$ being irrelevant, i.e., $q$ is not used in creating the expression tree. In the ideal case, all the irrelevant quantities are correctly predicted with high confidence, resulting in a large value for the sum of $\phi(q)$. The other term, denoted by $\phi(op)$, is the likelihood of selecting $op$ as the operator for an internal tree node. With these two factors, $Score(E)$ is formally defined as
\begin{equation} \label{eq:exp-tree-score}
Score(E) = w_1\sum_{q\in \mathcal{I}(E)}\phi(q) + \sum_{op \in\mathcal{N}}\phi(op)
\end{equation} 
where $\mathcal{I}(E)$ is the group of irrelevant quantities that are not included in expression $E$, and  $\mathcal{N}$ refers to the set of internal tree nodes. To further reduce the tree enumeration space, beam search is applied in~\cite{DBLP:conf/emnlp/RoyR15}. To generate the next state $T'$ from the current partial tree, the algorithm avoids choosing all the possible pairs of terms and determining their operator. Instead, only top-$k$ candidates with the highest partial scores are retained. Experimental results with $k=200$ show that the strategy achieves a good balance between accuracy and running time. The service is also published as a web tool~\cite{Roy16} and it can respond promptly to a math word problem.

ALGES~\cite{DBLP:journals/tacl/Koncel-Kedziorski15} differs from~\cite{DBLP:conf/emnlp/RoyR15} in two major ways. First, it adopts a more brutal-force manner to exploit all the possible equation trees. More specifically, ALGES does not discard irrevalent quantities, but enumerates all the syntactically valid trees. Integer Linear Programming (ILP) is applied as it can help enforce the constraints such as syntactic validity, type consistence and domain specific simplicity considerations. Consequently, its computation cost is dozens of times higher than that in~\cite{DBLP:conf/emnlp/RoyR15}, according to an efficiency evaluation in~\cite{DBLP:conf/aaai/Wang18}. Second, its scoring function is different from Equation~\ref{eq:exp-tree-score}. There is no need for the term $\phi(q)$ because ALGES does not build a classifier to check the quantity relevance. Besides the monotonic aggregation of the likelihood from local operator classifiers, the scoring function incorporates a new term $\phi(P)=\theta^T f_p$ to assign a coherence score for the tree instance. Here, $f_P$ is the global feature extracted from a problem text $P$, and $\theta$ refers to the parameter vector. 

The goal of ~\cite{DBLP:conf/emnlp/RoyUR16} is also to build an equation tree by parsing the problem text. It makes two assumptions that can simplify the tree construction, but also limit its applicability. First, the final output equation form is restricted to have at most two variables. Second, each quantity mentioned in the sentence can be used at most once in the final equation. The tree construction procedure consists of a pipeline of predictors that identify irrelevant quantities, recognize grounded variables, and generate the final equation tree. With customized feature selection and SVM based classifier, the relevant quantities and variables are extracted and used as the leaf nodes of the equation tree. The tree is built in a bottom-up manner. It is worth noting that to reduce the search space and simplify the tree construction, only adjacent nodes are combined to generate their parent node.

UnitDep~\cite{DBLP:conf/aaai/RoyR17} can be viewed as an extension work of~\cite{DBLP:conf/emnlp/RoyR15} by the same authors. An important concept, named Unit Dependency Graph (UDG), is proposed to enhance the scoring function. The vertices in UDG consist of the extracted quantities. If the quantity correspond to a rate (e.g., $8$ dollars per hour), the vertex is marked as RATE.  There are six types of edge relations to be considered, such as whether two quantities are associated with the same unit. Building the UDG requires additional annotation overhead as we need to train two classifiers for the nodes and edges. The node classifier determines whether a node is associated with a rate. The edge classifier predicts the type of relationship between any pair of quantity nodes. Given a valid unit dependency graph $G$ generated by the classifiers, its likelihood is defined as
\begin{equation}
\phi(G)=\sum_{v\in G \wedge LABEL(v)=rate} P(v) + \lambda \sum_{e\in G} P(e)
\end{equation}
In other words, we sum up the prediction probability for the RATE nodes and all the edges.  The new scoring function for an expression tree extends Equation~\ref{eq:exp-tree-score} by incorporating $\phi(G)$. Rules are defined to enforce the rate consistence between an expression tree $T$ and a candidate graph $G$. For example, if $v_i$ is the only node in the tree that is labeled RATE and it appears in the question, there should not exist a path from the leaf node to the root which only contains operators of addition and subtraction. Finally, the candidate graph $G$ with the highest likelihood and rate-consistent with $T$ is used to calculate the total score of $T$.

\subsection{DL-based Methods}
In recent years, deep learning (DL) has witnessed great success in a wide spectrum of ``smart'' applications, such as visual question answering~\cite{DBLP:conf/iccv/AntolALMBZP15, ZHANG2019268}, video captioning~\cite{song2018stochasticrnn,yi2018describing,yang2018adversial,DBLP:journals/tmm/GaoGZXS17}, personal interest inference~\cite{DBLP:journals/tois/GuoZWWCT18,DBLP:conf/icde/WuRYCZZ16,DBLP:journals/tois/WangNSZC17} and  smart transportation~\cite{WANG2019144}. The main advantage is that with sufficient amount of training data, DL is able to learn an effective feature representation in a data-driven manner without human intervention. It is not surprising to notice that several efforts have been attempted to apply DL for math word problem solving. Deep Neural Solver (DNS)~\cite{DBLP:conf/emnlp/WangLS17} is a pioneering work designed for equation set problems, which will be introduced in more details in the next section. Following DNS, there have emerged multiple DL-based solvers for arithmetic word problems. Seq2SeqET~\cite{DBLP:conf/emnlp/WangWCZL18} extended the idea of DNS by using expression tree as the output sequence. In other words, it applied seq2seq model to convert the problem text into an expression tree, which can be viewed as a template. Given the output of an expression tree or template, we can easily infer the numeric answer. To reduce the template space, equation normalization was proposed in Seq2SeqET so that duplicate representation of expression trees can be unified. StackDecoder~\cite{DBLP:journals/corr/abs-1811-00720} is also based on seq2seq model. Its encoder extracts semantic meanings of quantities in the question text and the decoder is equipped with a stack to facilitate tracking the semantic meanings of operands.  T-RNN~\cite{DBLP:conf/aaai/Wang19} can be viewed as an improvement of Seq2SeqET, in terms of quantity encoding, template representation and tree construction. First, an effective embedding network  with Bi-LSTM and self attention is used to vectorize the quantities. Second, the detailed operators in the templates are encapsulated to further reduce the number of template space.  For example, $n_1+n_2$, $n_1-n_2$, $n_1\times n_2$, and $n_1\div n_2$ are mapped to the same template $n_1\langle op\rangle n_2$. Third, a recursive neural network is applied to infer the unknown variables in the expression tree in a recursive manner.

Wang et al. made the first attempt of applying deep reinforcement learning to solve arithmetic word problems~\cite{DBLP:conf/aaai/Wang18}. The motivation is that deep Q-network has witnessed success in solving various problems with big search space such as playing text-based games~\cite{DBLP:conf/emnlp/NarasimhanKB15}, information extraction~\cite{DBLP:conf/emnlp/NarasimhanYB16}, text generation~\cite{2015arXiv151009202G} and object detection in images~\cite{DBLP:conf/iccv/CaicedoL15}. To fit the math problem scenario, they formulate the expression tree construction as a Markov Decision Process and propose the MathDQN that is customized from the general deep reinforcement learning framework. Technically, they tailor the definitions of states, actions, and reward functions which are key components in the reinforcement learning framework. By using a two-layer feed-forward neural network as the deep Q-network to approximate the Q-value function, the framework learns model parameters from the reward feedback of the environment. Compared to the aforementioned approaches, MathDQN iteratively picks the best operator for two selected quantities. This procedure can be viewed as beam search with $k=1$ when exploiting candidate expression trees. Its deep Q-network acts as the operator classifier and guides the model to select the most promising operator for tree construction.

\subsection{Dataset Repository and Performance Analysis}
The accuracy of arithmetic word problems is evaluated on the datasets that are manually harvested and annotated from online websites. These datasets are small-scale and contain hundreds of math problems. In this subsection, we make a summary on the datasets that have been used in the aforementioned papers. Moreover, we organize the performance results on these datasets into one unified table. We also make our best efforts to conduct additional experiments. The new results are highlighted in blue color. In this way, readers can easily identify the best performers in each dataset.

\subsubsection{Datasets}
There have been a number of datasets collected for the arithmetic word problems. We present their descriptions in the following and summarize the statistics of the datasets in Table~\ref{tbl:exp-arith-datasets}.
\begin{enumerate}
\item \textbf{AI2}~\cite{DBLP:conf/emnlp/HosseiniHEK14}. There are $395$ single-step or multi-step arithmetic word problems for the third, fourth, and fifth graders. It involves problems that can be solved with only addition and subtraction. The dataset is harvested from two websites: \texttt{math-aids.com} and \texttt{ixl.com} and comprises three subsets: MA1 (from \texttt{math-aids.com}), IXL (from \texttt{ixl.com}) and MA2 (from \texttt{math-aids.com}).  Among them, IXL and MA2 are more challenging than MA1 because IXL contains more information gaps and MA2 includes more irrelevant information in its math problems.

\item \textbf{IL}~\cite{DBLP:conf/emnlp/RoyR15}. The problems are collected from websites \texttt{k5learning.com} and \texttt{dadsworksheets.com}. The problems that require background knowledge (e.g., ``\textit{apple is fruit}'' and ``\textit{a week comprises $7$ days}'') are pruned. To improve the diversity, the problems are clustered by textual similarity. For each cluster, at most $5$ problems are retained. Finally, the dataset contains $562$ single-step word problems with only one operator, including addition, subtraction, multiplication, and division. 
%Each math problem also contains irrelevant quantities. %Again, we follow the same setting in ~\cite{DBLP:conf/emnlp/RoyR15} and perform a $5$-fold cross validation for this dataset.

\item \textbf{CC}~\cite{DBLP:conf/emnlp/RoyR15}. The dataset is designed for mult-step math problems. It contains $600$ multi-step problems without irrelevant quantities, harvested from \texttt{commoncoresheets.com}. The dataset involves various combinations of four basic operators, including (a) addition followed by subtraction; (b) subtraction followed by addition; (c) addition and multiplication; (d) addition and division; (e) subtraction and multiplication; and (f) subtraction and division. It is worth noting that this dataset does not incorporate irrelevant quantities in the problem text. Hence, there is no need to apply the quantity relevance classifier for the algorithms containing this component.

 %Following~\cite{DBLP:conf/emnlp/RoyR15}, we perform $6$-fold cross validation.

\item \textbf{SingleEQ}~\cite{DBLP:journals/tacl/Koncel-Kedziorski15}. The dataset contains both single-step and multi-step arithmetic problems and is a mixture of problems from a number of sources, including  \texttt{math-aids.com}, \texttt{k5learning.com}, \texttt{ixl.com} and a subset of the data from \textbf{AI2}. Each problem involves operators of multiplication, division, subtraction, and addition over non-negative rational numbers.

\item \textbf{AllArith}~\cite{DBLP:conf/aaai/RoyR17}. The dataset is  a mixture of  the data from \textbf{AI2}, \textbf{IL}, \textbf{CC} and \textbf{SingleEQ}. All mentions of quantities are normalized into digit representation. To capture how well the automatic solvers can distinguish between different problem types,  near-duplicate problems  (with over $80\%$ match of unigrams and bigrams) are removed. Finally, there remain $831$ math problems.

\item \textbf{MAWPS}~\cite{DBLP:conf/naacl/Koncel-Kedziorski16} is another testbed for arithmetic word problems with one unknown variable in the question. Its objective is to compile a dataset of varying complexity from different websites. Operationally, it combines the published word problem datasets used in~\cite{DBLP:conf/emnlp/HosseiniHEK14,DBLP:conf/acl/KushmanZBA14,DBLP:journals/tacl/Koncel-Kedziorski15,DBLP:conf/emnlp/RoyR15}. There are $2,373$ questions in the harvested dataset.

\item \textbf{Dolphin-S}. This is a subset of Dolphin18K~\cite{DBLP:conf/acl/HuangSLYM16} which originally contains $18,460$ problems and $5,871$ templates with one or multiple equations. The problems whose template is associated with only one problem are extracted as the dataset of \textbf{Dolphin-S}. It contains $115$ problems with single operator and $6,955$ problems with multiple operators.

\item \textbf{Math23K}~\cite{DBLP:conf/emnlp/WangLS17}. The dataset contains Chinese math word problems for elementary school students and is crawled from multiple online education websites. Initially, $60,000$ problems with only one unknown variable are collected. The equation templates are extracted in a rule-based manner. To ensure high precision, a large number of problems that do not fit the rules are discarded. Finally, $23,161$ math problems with $2,187$ templates remain.
\end{enumerate}

\begin{table}[t]
\begin{center}
\caption{Statistics of arithmetic word problem datasets.}\label{tbl:exp-arith-datasets}
\begin{tabular}{|c|c|c|c|c|c|c|c|c|c|}\hline
Dataset & $\#$ problems & $\#$ single-op & $\#$ multi-op & operators $\mathcal{O}$ \\ \hline\hline 
\textbf{MA1} &134 & 112 & 22 & $\{+,-\}$ \\ \hline
\textbf{IXL} & 140 & 119 & 21 & $\{+,-\}$\\ \hline
\textbf{MA2} & 121 & 96 & 25 &  $\{+,-\}$ \\ \hline
\textbf{AI2} &395 & 327 & 68 & $\{+,-\}$ \\ \hline
\textbf{IL} & 562 & 562 & 0 & $\{+,-,\times,\div\}$\\ \hline
\textbf{CC} & 600 & 0 & 600 &  $\{+,-,\times,\div \}$ \\ \hline
\textbf{SingleEQ} & 508 & 390 & 118 & $\{+,-,\times,\div\}$ \\ \hline
\textbf{AllArith} & 831 & 634 & 197 & $\{+,-,\times,\div\}$ \\ \hline
\textbf{MAWPS-S} & 2,373 & 1,311 & 1,062 & $\{+,-,\times,\div\}$ \\ \hline
\textbf{Dolphin-S} & 7,070  & 115  & 6,955  & $\{+,-,\times,\div\}$ \\ \hline
\textbf{Math23K} & 23,162 &  3,131 & 20,031  & $\{+,-,\times,\div\}$ \\ \hline
\end{tabular}
\end{center}
\end{table}

\begin{table*}[h]
\begin{center}
\caption{Accuracy of statistic-based and tree-based methods in solving arithmetic problems.}\label{tbl:exp-arithmetic}
\scalebox{0.92}{
\begin{tabular}{|c|c|c|c|c|c|c|c||c|c|c|}\hline
\multicolumn{3}{|c|}{} &  \textbf{AI2} & \textbf{IL} & \textbf{CC} &\textbf{SingleEQ} & \textbf{AllArith} & \textbf{Dolphin-S} & \textbf{MAWPS-S} & \textbf{Math23K} \\ \hline\hline 

\multicolumn{3}{|c|}{$\#$ problems}&  395 & 562 & 600 & 508 & 831 & 7,070 & 2,373 & 23,162 \\ \hline
\multicolumn{3}{|c|}{operators $\mathcal{O}$}&  $\{+,-\}$ & $\{+,-,\times,\div\}$ & $\{+,-,\times,\div\}$ & $\{+,-,\times,\div\}$ & $\{+,-,\times,\div\}$ & $\{+,-,\times,\div\}$ & $\{+,-,\times,\div\}$ & $\{+,-,\times,\div\}$\\ \hline\hline

\multirow{4}{*}{Statistic-based} & ARIS~\cite{DBLP:conf/emnlp/HosseiniHEK14} & 2014  & 77.7 & - & - & 48 &-&- &-&-\\ \cline{2-11}
 & Schema~\cite{sundaram-khemani:2015:W15-59} & 2015 & \textbf{88.64} & - & - &-&-&-&-&-\\ \cline{2-11}
 & Formula~\cite{DBLP:conf/acl/MitraB16} & 2016  & 86.07 & - & - &-&-&-&-&-\\ \cline{2-11}
& LogicForm~\cite{Liang16,DBLP:conf/ijcai/LiangHHLMS16} & 2016 & 84.8 & \textbf{80.1} & 53.5 &- &-&-&-&-\\  \hline \hline
\multirow{3}{*}{Tree-based} & ALGES~\cite{DBLP:journals/tacl/Koncel-Kedziorski15}& 2015  & 52.4 & 72.9 & 65 & 72 & 60.4&-&-&-\\ \cline{2-11}
& ExpressionTree~\cite{DBLP:conf/emnlp/RoyR15}& 2015 & 72	& 73.9 & 45.2 & {\color{blue}{66.38}} & 79.4&{\color{blue}{26.11}}&-&-\\ \cline{2-11}
 & UNITDEP~\cite{DBLP:conf/aaai/RoyR17} &2017  & 56.2 & 71.0 & 53.5 & \textbf{72.25} &\textbf{ 81.7 }&{\color{blue}{28.78}}&-&-\\ \hline
\multirow{4}{*}{DL-based} & MathDQN~\cite{DBLP:conf/aaai/Wang18} & 2018 & 78.5 & 73.3 &\textbf{75.5} & {\color{blue}{52.96}} & {\color{blue}{72.68}}&{\color{blue}{30.06}} & 60.25&-\\ \cline{2-11}
 & Seq2SeqET~\cite{DBLP:conf/emnlp/WangWCZL18} & 2018 & - & - &- & - & - &-&-& 66.7\\ \cline{2-11}
 & T-RNN~\cite{DBLP:conf/aaai/Wang19} & 2019 & - & - &- & - & - &\color{blue}{39.1}& 66.8& 66.9\\ \cline{2-11}
 & StackDecoder~\cite{DBLP:journals/corr/abs-1811-00720} & 2019 & - & - &- & - & - &-& -&65.8\\ \cline{2-11}
 \hline
% KAZB~\cite{DBLP:conf/acl/KushmanZBA14}& 2014 & 89.6 & 51.1 & 51.2 & 64.0 & 73.7 & 2.3 & 0.48 & 73.7 \\\hline 
\end{tabular}
}
\end{center}
\end{table*}
\subsubsection{Performance Analysis}
Given the aforementioned datasets, we merge the experimental results reported from previous works into one table. Such a unified organization can facilitate readers in identifying the methods with superior performance in each dataset. As shown in Table~\ref{tbl:exp-arithmetic}, the rows refer to the corpus of datasets and the columns are the statistic-based and tree-based methods. The cells are filled with the accuracies of these algorithms when solving math word problems in different datasets. We conduct additional experiments to cover all the cells by the tree-based solutions. These new experiment results are highlighted in blue color. Those with missing value are indicated by ``-'' and it means that there was no experiment conducted for the algorithm in the particular dataset. The main reason is that they require particular efforts on logic templates and annotations, which are very non-trivial and cumbersome for experiment reproduction. There is no algorithm comparison for the dataset \textbf{Math23K} because the problem text is in Chinese and the feature extraction technologies proposed in the statistic-based and tree-based approaches are not applicable.
%Two of Roy's approaches, ~\cite{DBLP:journals/tacl/RoyVR15} and~\cite{DBLP:conf/emnlp/RoyUR16}, are not included in the table because they have not been compared with any other existing arithmetic problem solvers. 
From the results in Table~\ref{tbl:exp-arithmetic}, we derive the following observations worth noting and provide reasonings to explain the results.

First, the statistic-based methods with advanced logic representation, such as Schema~\cite{sundaram-khemani:2015:W15-59}, Formula~\cite{DBLP:conf/acl/MitraB16} and LogicForm~\cite{Liang16,DBLP:conf/ijcai/LiangHHLMS16}, achieve dominating performance in the \textbf{AI2} dataset. Their superiority is primarily owned to the additional efforts on annotating the text problem with more advanced logic representation. These annotations allow them to conduct fine-grained reasoning. In contrast, ARIS~\cite{DBLP:conf/emnlp/HosseiniHEK14} does not work as good because it focuses on ``change'' schema of quantities and does not fully exploit other schemas like ``compare''~\cite{sundaram-khemani:2015:W15-59}.  Since there are only hundreds of math problems in the datasets, it is feasible to make an exhaustive scan on the math problems and manually define the templates to fit these datasets. For instance, all quantities and the main-goal are first identified by rules in LogicForm~\cite{Liang16,DBLP:conf/ijcai/LiangHHLMS16} and explicitly associated with their role-tags.  Thus, with sufficient human intervention, the accuracy of statistic-baesd methods in \textbf{AI2} can boost to $88.64\%$, much higher than that of tree-based methods. Nevertheless, these statistic-based methods are considered as brittle and rigid~\cite{DBLP:conf/acl/HuangSLYM16} and not scalable to handle large and diversified datasets, primarily due to the heavy annotation cost to train an accurate mapping between the text and the logic representation.

Second, the results of tree-based methods in \textbf{AI2}, \textbf{IL} and \textbf{CC} are collected from~\cite{DBLP:conf/aaai/Wang18} where the same experimental setting of $3$-fold cross validation is applied. It is interesting to observe that ALGES~\cite{DBLP:journals/tacl/Koncel-Kedziorski15}, ExpressionTree~\cite{DBLP:conf/emnlp/RoyR15} and UNITDEP~\cite{DBLP:conf/aaai/RoyR17} cannot perform equally well on the three datasets. ALGES works poorly in \textbf{AI2} because irrelevant quantities exist in its math problems and ALGES is not trained with a classifier to get rid of them. However, it outperforms ExpressionTree and UNITDEP by a wide margin in the \textbf{CC} dataset because \textbf{CC} does not involve irrelevant quantities. In addition, this dataset only contains multi-step math problems. ALGES exploits the whole search space to enumerate all the possible trees, whereas ExpressionTree and ALGES use beam search for efficiency concern. UNITDEP does not work well in \textbf{AI2} because this dataset only involves operators $+$ and $-$ and the unit dependency graph does not take effect. Moreover, its proposed context feature poses a negative impact on the \textbf{AI2} dataset. After removing this feature from the input vector fed to the classifier, the accuracy in \textbf{AI2} increases from $56.2\%$ to $74.7\%$, but the result in \textbf{CC} drops from the current accuracy of $53.5\%$ to $47.3\%$. Such an observation implies the limitation of hand-crafted features in UNITDEP. Among the three datasets \textbf{AI2}, \textbf{IL} and \textbf{CC}, MathDQN~\cite{DBLP:conf/aaai/Wang18} achieves leading or comparable performance. In the \textbf{CC} dataset, which contains only multi-step problems and is considered as the most challenging one, MathDQN yields remarkable improvement and boosts the accuracy from $65\%$ (derived by ALGES) to $75.5\%$. This is because MathDQN models the tree construction as Markov Decision Process and leverage the strengths of deep Q-network (DQN). By using a two-layer feed-forward neural network as the deep Q-network to approximate the Q-value function, the framework learns model parameters from the reward feedback of the environment. Consequently, the RL framework demonstrates higher generality and robustness than the other tree-based methods when handling complicated scenarios. In the \textbf{IL} dataset, its performance is not superior to ExpressionTree as \textbf{IL} only contains one-step math problems. There is no need for hierarchical tree construction and cannot expose the strength of Markov Decision Process in MathDQN or the exhaustive enumeration strategy in ALGES.

As to the datasets of SingleEQ and AllArith, UNITDEP is a winner in both datasets, owning to the effectiveness of the proposed unit dependency graph (UDG). In the math problems with operators $\{\times,\div\}$, the unit and rate are important clues to determine the correct quantities and operators in the math expression. The UDG poses constraints on unit compatibility to filter the false candidates in the expression tree construction. It can alleviate the brittleness of the unit extraction system, even though it requires additional annotation overhead in order to induce UDGs.

%In the \textbf{SingleEQ} dataset, ALGES is compared with ARIS and shows a dominating superiority. 
%However, this is an unfair comparison as ARIS cannot handle the problems with multiplication and division.

%However, these methods are rigid and not scalable to handle large datasets because they require considerable annotation overhead to train an accurate mapping between the text and the logic representation. Their performances on the IL and CC datasets have not been evaluated by the researchers in the community, partly due to the heavy annotation cost. 
%Hereby, there is no clear evidence to support their generality and there is a possibility that the manually defined templates may overfit the problems in \textbf{AI2}. 

Last but not the least, these experiments were conducted on small-scale datasets and their performances on larger and more diversified datasets remain unclear. Recently, Huang et al. have noticed the gap and released Dolphin18K~\cite{DBLP:conf/acl/HuangSLYM16} which contains $18,460$ problems and $5,871$ templates with one or multiple equations. The findings in ~\cite{DBLP:conf/acl/HuangSLYM16} are astonishing. The accuracies of existing approaches for equation set problems, which will be introduced in the next section, degrade sharply to less than $25\%$. These methods cannot even perform better than a simple baseline that first uses text similarity to find the most similar text problem in the training dataset and then fills the number slots in its associated equation template. Math23K is another large-scale dataset which contains Chinese math word problems. The results are shown in the last three columns of Table~\ref{tbl:exp-arithmetic}. We can see that T-RNN is state-of-the-art method in the two largest datasets \textbf{Dolphin-S} and \textbf{Math23K}. 

%To discover the performance of existing arithmetic problem solvers in large-scale datasets, we conduct similar experiments to examine the performance of tree-based approaches in \textbf{Dolphin-S}.  The results are shown in the last two columns of Table~\ref{tbl:exp-arithmetic}, leading to two conclusions. First, the overall performances of existing tree-based arithmetic problem solvers are not promising in large-scale and diversified datasets. There is still great room for improvement in the research area of arithmetic math word problem solver. Second, MathDQN exhibits superior performance over its two tree-based competitors, verifying its robustness by using the deep reinforcement learning framework.

\section{Equation Set Solver}\label{sec:eqn-set}
The equation set problems are much more challenging because they involve multiple unknown variables to resolve and require to formulate a set of equations to obtain the final solution. The aforementioned arithmetic math problem can be viewed as a simplified variant of equation set problem with only one unknown variable. Hence, the methods introduced in this section can also be applied to solve the problems in Section~\ref{sec:arithmetic}.

\begin{figure}[h]
\centering
\includegraphics[width=70mm, keepaspectratio]{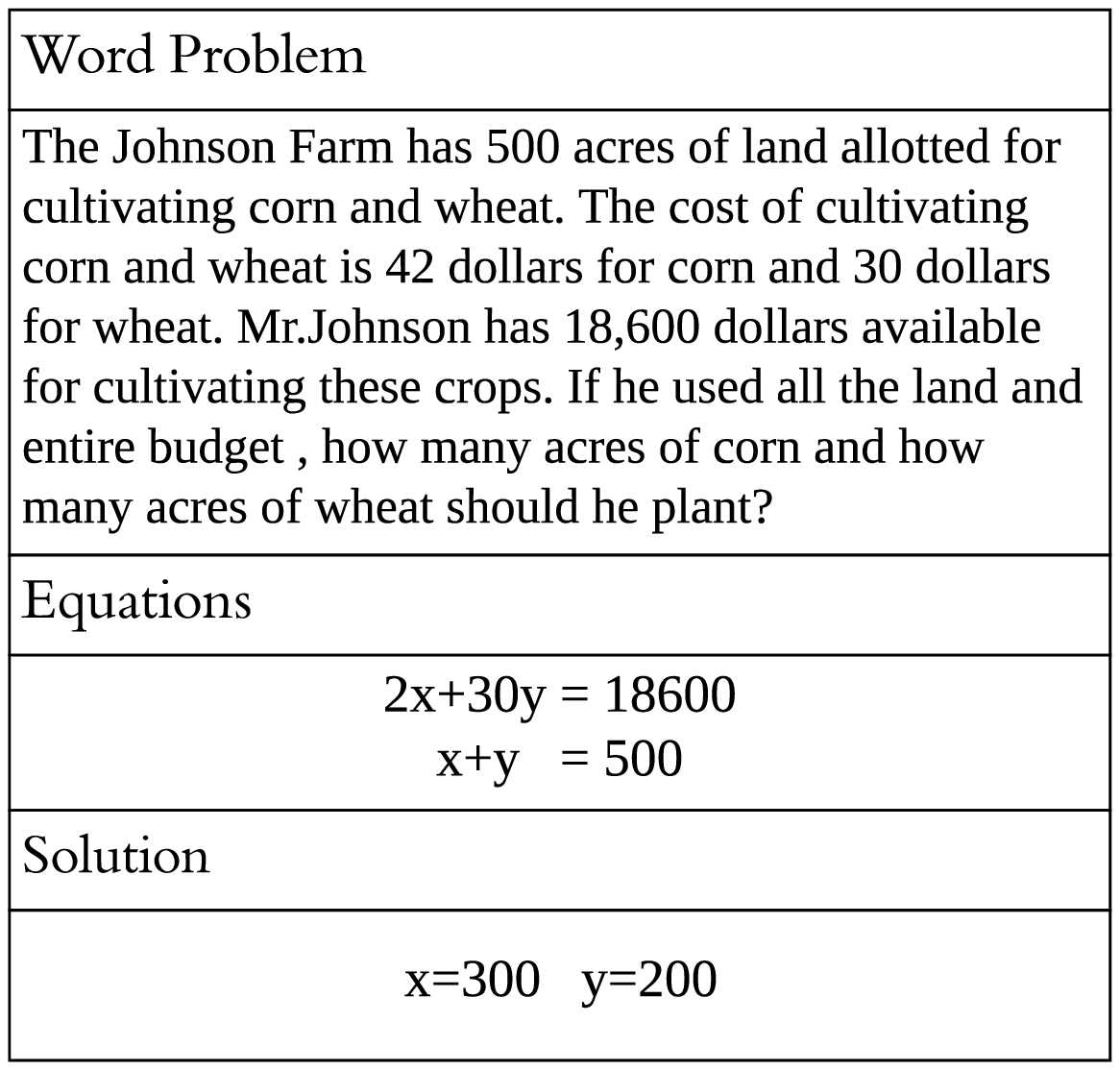} 
\caption{An example of equation set problem.}
\label{fig:example-eq-set}
\end{figure}

Figure~\ref{fig:example-eq-set} shows an example of equation set problem. There are two unknown variables, including the acres of corn and the acres of wheat, to be inferred from the text description. A standard solution to this problem is to use variables $x$ and $y$ to represent the number of corn and wheat, respectively. From the text understanding, we can formulate two equations $42x+30y=18600$ and $x+y=500$. Finally, the values of $x$ and $y$ can be inferred.

Compared to the arithmetic problem, the equation set problem contains more numbers of unknown variables and numbers in the text, resulting in a much larger search space to enumerate valid candidate equations. Hence, the methods designed for arithmetic problems can be hardly applied to solve equation set problems. For instance, the tree-based methods assume that the objective is to construct a single tree to maximize a scoring function. They require substantial revision to adjust the objective to building multiple trees which has exponentially higher search space. This would be likely to degrade the performance. In the following, we will review the existing methods, categorize them into four groups from the technical perspective, and examine how they overcome the challenge.

%In the following, XXX. Similarity-based method requires a very large dataset to achieve good accuracy. Template-based method is the mainstream algorithm.

\subsection{Parsing-Based Methods}
The work of \cite{DBLP:conf/emnlp/ShiWLLR15} can be viewed as an extension of tree-based approaches to solve a math problem with multiple equations. Since the objective is no longer to build an equation tree, a meaning representation language called \textit{DOL} is designed as the structural semantic representation of natural language text. The core component is a semantic parser that transforms the textual sentences into DOL trees. The parsing algorithm is based on context-free grammar (CFG)~\cite{journals/mst/Knuth68,DBLP:journals/cacm/Earley70}, a popular mathematical system for modeling constituent structure in natural languages. For every DOL node type, the lexicon and grammar rules are constructed in a semi-supervised manner. The association between math-related concepts and their grammar rules is manually constructed. Finally, the CFG parser is built on top of $9,600$ grammar rules. During the parsing, a score is calculated for each DOL node and the derivation of the DOL trees with the highest score is selected to obtain the answer via a reasoning module.

\subsection{Similarity-Based Methods}\label{sec:similarity}
The work of~\cite{DBLP:conf/acl/HuangSLYM16} plays an important role in the research line of automatic math problem solvers because it rectifies the understanding of technology development in this area. It, for the first time, examines the performance of previous approaches in a large and diversified dataset and derives astonishing experimental findings. The methods that claimed to achieve an accuracy higher than $70\%$ in a small-scale and self-collected dataset exhibit very poor performance in the new dataset. In other words, none of the methods proposed before~\cite{DBLP:conf/acl/HuangSLYM16} is really general and robust. Hence, the authors reach a conclusion that the math word problems still have much room for improvement.

A new baseline method based on text similarity, named SIM, is proposed in~\cite{DBLP:conf/acl/HuangSLYM16}. In the first step, the problem text is converted into a word vector whose values are the associated TF-IDF scores~\cite{DBLP:journals/tois/ZhangNLTCS17}. The similarity between two problems (one is the query problem to solve and the other is a candidate in the training dataset with known solutions) is calculated by the Jaccard similarity between their converted vectors. The problem with the highest similarity score is identified and its equation template is used to help solve the query math problem. In the second step, the unknown slots in the template are filled. With the availability of a large training dataset, the number filling is conducted in a simple and effective way. It finds an instance in the training dataset associated with the same target template and the minimum edit-distance to the query problem, and aligns the numbers in these two problems with ordered and one-to-one mapping. It is considered a failure if these two problems do not contain the same number of quantities.

\subsection{Template Based Methods}
There are two methods~\cite{DBLP:conf/acl/KushmanZBA14, DBLP:conf/emnlp/ZhouDC15} that pre-define a collection of equation set templates. Each template contains a set of\textit{ number slots} and \textit{unknown slots}. The \textit{number slots} are filled by the numbers extracted from the text and the \textit{unknown slots} are aligned to the nouns. An example of template may look like
\begin{eqnarray*}
u_1+u_2-n_1 &=& 0 \\
n_2\times u_1 + n_3\times u_2 - n_4 &=&  0
\end{eqnarray*}
where $n_i$ is a number slot and $u_i$ represents an unknown variable.

To solve an equation set problem, these approaches first identify a candidate template from the corpus of pre-defined templates. The next task is to fill the \textit{number slots} and \textit{unknown slots} with the information extracted from the text. 
Finally, the instance with the highest probability is returned. This step often involves a scoring function or a rank-aware classifier such as RankSVM~\cite{herbrich2000large}. A widely-adopted practice is to define the probability of each instance of derivation $y$ based on the feature representation $x$ for a text problem and a parameter vector $\theta$, as in~\cite{DBLP:conf/emnlp/RoyUR16,DBLP:conf/acl/KushmanZBA14, DBLP:conf/emnlp/ZhouDC15}:
$$p(y|x;\theta)=\frac{e^{\theta\cdot \phi(x,y)}}{\sum_{y'\in \mathcal{Y}}e^{\theta\cdot \phi(x,y')}}$$ 
With the optimal derivation instance $y^{opt}$, we can obtain the final solution.

The objective of the work from Kushman et al.~\cite{DBLP:conf/acl/KushmanZBA14} is to maximize the total probabilities of $y$ that leads to the correct answer. The latent variables $\theta$ are learned by directly optimizing the marginal data log-likelihood. More specifically, L-BFGS~\cite{Nocedal2006NO} is used to optimize the parameters. The search space is exponential to the number of slots because each number in the text can be mapped to any \textit{number slot} and the nouns are also candidates for the \textit{unknown slots}. In practice, the search space is too huge to find the optimal $\theta$ and beam search inference procedure is adopted to prevent enumerating all the possible $y$ leading to the correct answer.  For the completion of each template, the next slot to be considered is selected according to a pre-defined canonicalized ordering and only top-$k$ partial derivations are maintained.

Zhou et al. proposed an enhanced algorithm for the template-based learning framework~\cite{DBLP:conf/emnlp/ZhouDC15}. First, they only consider assigning the number slots with numbers extracted from the text. The underlying logic is that when the number slots have been processed, it would be an easy task to fill the unknown slots. In this way, the hypothesis space can be significantly reduced. Second, the authors argue that the beam search used in~\cite{DBLP:conf/acl/KushmanZBA14} does not exploit all the training samples, and its resulting model may be sub-optimal. To resolve the issue, the max-margin objective~\cite{vapnik2013nature} is used to train the log-linear model. The training process is turned into a QP problem that can be efficiently solved with the constraint generation algorithm~\cite{Koller:2009:PGM:1795555}.

Since the annotation of equation templates is expensive, a key challenge to KAZB and ZDC is the lack of sufficient annotated data. To resolve the issue,  Upadhyay et al. attempted to exploit the large number of algebra word problems that have been posted and discussed in online forums~\cite{DBLP:conf/emnlp/UpadhyayCCY16}. These data are not explicitly annotated with equation templates but their numeric answers are extracted with little or no manual effort. The goal of ~\cite{DBLP:conf/emnlp/UpadhyayCCY16} is to improve a strong solver trained by fully annotated data with a large number of math problems with noisy and implicit supervision signals. The proposed \textit{MixedSP} algorithm makes use of both explicit and implicit supervised examples mixed at the training stage and learns the parameters jointly. With the learned model to formulate the mapping between an algebra word problem and an equation template, the math problem solving strategy is similar to KAZB and ZDC. All the templates in the training set have to be exploited to find the best alignment strategy.

The aforementioned template-based methods suffer from two drawbacks~\cite{DBLP:conf/emnlp/HuangSYL17}. First, the math concept is expressed as an entire template and may fail to work well when the training instances are sparse. Second, the learning process relies on lexical and syntactic features such as the dependency path between two slots in a template. Such a huge and sparse feature space may play a negative impact on effective feature learning. Based on these two arguments, FG-Expression~\cite{DBLP:conf/emnlp/HuangSYL17} parses an equation template into fine-grained units, called \textit{template fragment}. Each template is represented in a tree structure as in Figure~\ref{fig:tree-examples} and each fragment represents a sub-tree rooted at an internal node. The main objective and challenge in~\cite{DBLP:conf/emnlp/HuangSYL17} are learning an accurate mapping between textual information and template fragments. For instance, a text piece ``\textit{20\% discount}'' can be mapped to a template fragment $1-n_1$ with $n_1=0.2$. Such mappings are extracted in a semi-supervised way from training datasets and stored as part of the sketch for templates. The proposed solution to a math problem consists of two stages. First, RankSVM model~\cite{herbrich2000large} is trained to select top-$k$ templates. The features used for the training incorporate textual features, quantity features and solution features. It is worth noting that the proposed template fragment is applied in the feature selection for the classifier. The textual features preserve one dimension to indicate whether the problem text contains textual expressions in each template fragment. In the second stage, the alignment is conducted for the $k$ templates and the one with the highest probability is used to solve the problem. The features and rank-based classifier used to select the best alignment are similar to those used in the first stage. Compared to the previous template-based methods, FG-Expression also significantly reduces the search space because only top-$k$ templates are examined whereas previous methods align numbers for all the templates in the training dataset.

\begin{table*}[t]
\begin{center}
\caption{Statistics of datasets for equation set problems.}\label{tbl:dataset-eqn-set}
\begin{tabular}{|c|c|c|c|c|c|c|c|}\hline
 Datasets & Proposed in & $\#$ of problems & $\#$ of templates & $\frac{\# \text{ of problems}}{\# \text{ of templates}}$ & $\#$ of single-eq problems & $\#$ of words & $\#$ of sentences \\ \hline\hline 
 \textbf{ALG514} &  KAZB~\cite{DBLP:conf/acl/KushmanZBA14} & 514 & 28  & 18.36 & 91 & 1.62k & 19.3k \\ \hline
 \textbf{Dolphin1878} & DOL~\cite{DBLP:conf/emnlp/ShiWLLR15}& 1,878 & 1,183 & 1.59& 712 & 3.30k & 41.4k \\ \hline
 \textbf{DRAW1K} &MixedSP~\cite{DBLP:conf/emnlp/UpadhyayCCY16} & 1,000 &232  & 4.31 & 255 & 1.38k & 13.8k\\ \hline
 \textbf{Dolphin18K} & SIM~\cite{DBLP:conf/acl/HuangSLYM16} & 18,460 & 5,871 &3.14 & 8,333 & 49.9k & 604k\\ \hline
\end{tabular}
\end{center}
\end{table*}

\subsection{DL-Based Methods}
%In recent years, deep learning (DL) has witnessed great success in a wide spectrum of ``smart'' applications, such as visual question answering~\cite{DBLP:conf/iccv/AntolALMBZP15, XXXX}, video captioning~\cite{song2018stochasticrnn,yi2018describing,yang2018adversial,DBLP:journals/tmm/GaoGZXS17}, personal interest inference~\cite{DBLP:journals/tois/GuoZWWCT18,DBLP:conf/icde/WuRYCZZ16,DBLP:journals/tois/WangNSZC17} and  smart transportation~\cite{WANG2019144}. The main advantage is that with sufficient amount of training data, DL is able to learn an effective feature representation in a data-driven manner without human intervention. It is not surprising to notice that several efforts have been attempted to apply DL for math word problem solving. 
Deep Neural Solver (DNS)~\cite{DBLP:conf/emnlp/WangLS17} is the first deep learning based algorithm that does not rely on hand-crafted features. This is a milestone contribution because all the previous methods (including MathDQN) require human intelligence to help extract features that are effective. The deep model used in DNS is a typical sequence to sequence (seq2seq) model~\cite{DBLP:conf/nips/SutskeverVL14,DBLP:journals/corr/LuongLSVK15,DBLP:conf/emnlp/WisemanR16}. The words in the problem are vectorized into features through word embedding techniques~\cite{DBLP:conf/nips/MikolovSCCD13,DBLP:conf/emnlp/PenningtonSM14}. In the encoding layer, GRU~\cite{DBLP:journals/corr/ChoMGBSB14} is used as the Recurrent Neural Network (RNN) to capture word dependency because compared to LSTM~\cite{DBLP:journals/neco/HochreiterS97}, GRU has fewer parameters in the model and is less likely to be over-fitting in small datasets. This seq2seq model translates math word problems to equation templates, followed by a number mapping step to fill the slots in the equation with the quantities extracted from the text. To ensure that the output equations by the model are syntactically correct, five rules are pre-defined as validity constraints. For example, if the $i^{th}$ character in the output sequence is an operator in $\{+,-,\times,\div\}$, then the model cannot result in $c\in\{+,-,\times,\div,),=\}$ for the $(i+1)^{th}$ character.

To further improve the accuracy, DNS enhances the model in two ways. First, it builds a LSTM-based binary classification model to determine whether a number is relevant. This is similar to the relevance model trained in ExpressionTree~\cite{DBLP:conf/emnlp/RoyR15} and  UNITDEP~\cite{DBLP:conf/aaai/RoyR17}. The difference is that DNS uses LSTM as the classifier with unsupervised word-embedding features whereas ExpressionTree and UNITDEP use SVM with hand-crafted features. Second, the seq2seq model is integrated with a similarity-based method~\cite{DBLP:conf/acl/HuangSLYM16} introduced in Section~\ref{sec:similarity}. Given a pre-defined threshold, the similarity-based retrieval strategy is selected as the solver if the maximal similarity score is higher than the threshold. Otherwise, the seq2seq model is used to solve the problem. Another follow-up of DNS was proposed recently in~\cite{2018arXiv180410718R}. Instead of using GRU and LSTM, the math solver examines the performance of other seq2seq models when applied in mapping the problem text to equation templates. In particular, two models including BiLSTM~\cite{DBLP:journals/neco/HochreiterS97} and structured self-attention~\cite{2017arXiv170303130L}, were examined respectively for the equation template classification task. Results show that both models achieve comparable performance. CASS~\cite{DBLP:conf/coling/HuangLLY18} is an extension of seq2seq model. It adjusts the output sequence generation process by incorporating the copy and alignment mechanism. Reinforcement learning (RL) technique is adopted and the model is trained using policy gradient. Results show that the model trained by RL is more accurate than using maximum likelihood estimation (MLE) as the objective function.

\begin{table*}[h]
\begin{center}
\caption{Accuracies of equation set problem solvers on existing datasets.}\label{tbl:exp-eqn-set}
\begin{tabular}{|c|c|c|c|c|c|c|}\hline
\multicolumn{3}{|c|}{} & \textbf{ALG514} & \textbf{Dolphin1878} & \textbf{DRAW1K} & \textbf{Dolphin18K} \\ \hline\hline 

\multicolumn{3}{|c|}{$\#$ of problems} & 514 & 1,878 & 1,000 & 18,460 \\ \hline
\multicolumn{3}{|c|}{$\#$ of templates} & 28 & 1,183 & 232 & 5,871 \\ \hline\hline 

Statistic-based & DOL~\cite{DBLP:conf/emnlp/ShiWLLR15} & 2015 & & 60.2 & &  \\ \hline
 Similarity-based  & SIM~\cite{DBLP:conf/acl/HuangSLYM16} & 2016 & 70.1 & 29 & {\color{blue}{25.5}} & 18.4\\ \hline
\multirow{4}{*}{Template-based} & KAZB~\cite{DBLP:conf/acl/KushmanZBA14} & 2014 & 68.7 & & 43.2  &  \\ \cline{2-7}
 & ZDC~\cite{DBLP:conf/emnlp/ZhouDC15} & 2015 & 79.7 & & & 17.9  \\ \cline{2-7}
 & MixedSP~\cite{DBLP:conf/emnlp/UpadhyayCCY16} & 2016 & 83.0 & & 59.5 & \\ \cline{2-7}
 & FG-Expression~\cite{DBLP:conf/emnlp/HuangSYL17} & 2017 & & & & 28.4 \\ \hline
%DL-based & DNS~\cite{DBLP:conf/emnlp/WangLS17} & 2017 & 70.1 &{\color{blue}{28.29}} & {\color{blue}{31}} & {\color{blue}{21.6}} \\ \hline
\multirow{2}{*}{DL-based} & DNS~\cite{DBLP:conf/emnlp/WangLS17} & 2017 & 70.1 &{\color{blue}{28.29}} & {\color{blue}{31}} & {\color{blue}{21.6}}\\  \cline{2-7}
&CASS~\cite{DBLP:conf/coling/HuangLLY18} & 2018 & 82.5 & 29.7 & - & 29.0\\  \hline
% KAZB~\cite{DBLP:conf/acl/KushmanZBA14}& 2014 & 89.6 & 51.1 & 51.2 & 64.0 & 73.7 & 2.3 & 0.48 & 73.7 \\\hline 
\end{tabular}
\end{center}
\end{table*}

\subsection{Dataset Repository and Performance Analysis}
Similar to the organization of Section~\ref{sec:arithmetic}, we summarize the dataset repository and performance analysis for the equation set solvers.

\subsubsection{Benchmark Datasets}
There have been four datasets specifically collected for the equation set problems that involve multiple unknown variables. We present their descriptions in the following and summarize the statistics of the datasets in Table~\ref{tbl:dataset-eqn-set}. We use $\frac{\# \text{ of problems}}{\# \text{ of templates}}$ to report the average number of problems associated with each template.  We noticed that in each dataset, a small fraction of problems are associated with one unknown variable in the template. Thus, we also report the number of single-equation problems in each dataset.
%Such statistics can reflect the difficulty level of the datasets.
\begin{enumerate}

\item \textbf{ALG514}~\cite{DBLP:conf/acl/KushmanZBA14}. The dataset is crawled from \texttt{Algebra.com}, a crowd-sourced tutoring website. The problems are posted by students. The problems with information gap or require explicit background knowledge are discarded. Consequently, a set of $1024$ questions is collected and cleaned by crowd-workers in Amazon Mechanical Turk. These problems are further filtered as the authors require each equation template to appear for at least $6$ times. Finally, $514$ problems are left in the dataset.

\item \textbf{Dolphin1878}~\cite{DBLP:conf/emnlp/ShiWLLR15}. Its math problems are crawled from two websites: \texttt{albegra.com} and \texttt{answers.yahoo.com}. 
%For math problems in \texttt{albegra.com}, they has $XXX$ percent overlap with \textbf{ALG514}. 
For math problems in \texttt{answers.yahoo.com}, the math equations and answers are manually added by human annotators. Finally, the dataset combined from the two sources contains $1,878$ math problems with $1183$ equation templates.

\item \textbf{DRAW1K}~\cite{DBLP:conf/eacl/UpadhyayC17}. The authors of \textbf{DRAW1K} argued that \textbf{Dolphin1878} has limited textual variations and lacks narrative. This motivated them to construct a new dataset that is diversified in both vocabularies and equation systems. With these two objectives, they constructed  \textbf{DRAW1K} with exactly $1,000$ linear equation problems that are crawled and filtered from \texttt{algebra.com}.

\item \textbf{Dolphin18K}~\cite{DBLP:conf/acl/HuangSLYM16}. The dataset is collected and rectified mainly from the math category of Yahoo! Answers\footnote{https://answers.yahoo.com/}. The problems, equation system annotations, and answers are extracted semi-automatically, with great intervention of human efforts. The procedure consists of four stages: removing irrelevant problems, cleaning problem text, extracting gold answers and constructing equation system annotations. The harvested dataset is so far the largest one, with $18,460$ problems and $5,871$ equation templates. Since the dataset is collected from online forum, there could exist errors in the annotations and answers. The quality may not as high as those crawled from professional websites or constructed by paid crowd workers. Nevertheless, the large volume and high diversify make Dolphin18K a challenging dataset that is useful to examine the generality and robustness of a MWP solver. It has become a popular dataset as a number of MWP solvers used it for experimental evaluation.

\item \textbf{AQuA}~\cite{DBLP:conf/naacl/Koncel-Kedziorski16} is a large-scale dataset published by DeepMind. The authors first collected 34,202 multi-choice math questions covering a broad range of topics and difficulty levels. The dataset is rather challenging because many problems are from GMAT (Graduate Management Admission Test) and GRE (General Test). In contrast, most of the aforementioned MWP datasets are targeted for elementary school students. Moreover, crowd workers are also involved in the dataset construction of AQuA, contributing 70,318 additional questions. In total, the dataset contains around $100,000$ annotated math problems.

\end{enumerate}

\subsubsection{Performance Analysis}
The performances of the equation set solvers on the existing datasets are reported in Table~\ref{tbl:exp-eqn-set}. From the experimental results, we derive the following observations and discussions.

First, there are many empty cells in the table. In the ideal case, the algorithms should be conducted on all the available benchmark datasets and compared with all the previous competitors.  The reasons are multi-fold, such as limitation of the implementation (e.g., as Upadhyay stated in~\cite{DBLP:conf/emnlp/UpadhyayCCY16}, they could not run ZDC on \textbf{DRAW1K} because ZDC can only handle limited types of equation systems), delayed release of the dataset (e.g., the \textbf{Dolphin18K} dataset has not been released when the work of DNS is published) or unfitness in certain scenarios (e.g., the experiments of FG-Expression~\cite{DBLP:conf/emnlp/HuangSYL17} were only conducted on \textbf{Dolphin18K} because the authors considered that the previous datasets are not suitable due to their limitation in scalability and diversity). It is noticeable that such an incomplete picture brings difficulty to judge the performance and may miss certain insightful findings.

Second, \textbf{ALG514} is the smallest dataset and also the most widely adopted dataset for performance evaluation. Among the template-based methods, MixedSP outperforms KAZB and ZDC because it benefits from the mined implicit supervision from an external source of additional $2,000$ samples. As reported in~\cite{DBLP:conf/emnlp/UpadhyayCCY16}, if only the explicit dataset (i.e., the problems in \textbf{ALG514}) is used, its performance is slightly inferior to ZDC. A possible reason to explain this is that ZDC uses a richer set of features based on POS tags, coreference and dependency parses. In contrast, MixedSP only uses features based on POS tags. It is also interesting to see that SIM and DNS obtain the same accuracy on \textbf{ALG514}. This is the dataset is too small to train an effective deep learning model. The reported accuracy of seq2seq model is only $16.1\%$ in \textbf{ALG514}. DNS is a hybrid approach that combines a seq2seq model and similarity retrieval model. It means the deep learning model does not take any effect when handling problems in \textbf{ALG514}. 

Finally, the datasets of \textbf{Dolphin1878} and \textbf{DRAW1K} are released for the approaches of DOL and MixedSP, respectively. In the experimental settings, simple baselines such as SIM based on textual similarity or KAZB which is the earliest template-based method, are selected. It is not surprising to see that \textbf{Dolphin1878} and \textbf{DRAW1K} outperform their competitors by a large margin. Nevertheless, the research for equation set solvers has shifted to proposing methods that can work well in a large and diversified dataset such as \textbf{Dolphin18K}. We implemented our own version of DNS and evaluated its performance on the large dataset. Unfortunately, we did not observe a higher accuracy derived from DNS. The reasons could be two-fold. First, our implementation may not be optimized and the model parameters may not be well tuned. Second, there are thousands of templates in the datasets, which may bring challenges for the classification task. CASS further improved the accuracy on \textbf{Dolphin18K} to $29\%$, verifying the effectiveness of their proposed copy and alignment mechanism, as well as the optimization based on policy gradient.

% Nevertheless, with the availability of large-scale datasets, applying deep learning models for MWPs is still an interesting research direction that deserves intensive attention. 

%In the future, we expect to see that algorithms proposed for arithmetic math problems are compared in \textbf{Math23K} and algorithms proposed for equation set solvers are compared in \textbf{Dolphin18K}.

%IN SMALL DATASET, MIXEDSP IS THE BEST?IN LARGER DAATASET, DNS CAN BE BETTER?

\eat{
\subsection{Comparison with Arithmetic Problem Solvers}

Compare similarities and differences between arithmetic word problem solvers and equation set problem solver. 

Tree is a strong connection.

Report the results of equation-set solvers on arithmetic datasets. Such as Math23K
}

\section{Feature Extraction}\label{sec:feature}
Before moving to the third class of math word problems (namely, geometry problems), we take a detour to discuss feature extraction, which plays a vital component in all systems (except DSN) discussed in sections 2 and 3. Effective feature construction, in an either supervised or semi-supervised manner, can significantly boost the accuracy. For instance, SIFT~\cite{DBLP:conf/iccv/Lowe99,DBLP:journals/ijcv/Lowe04} and other local descriptors~\cite{DBLP:conf/cvpr/DalalT05,DBLP:journals/cviu/BayETG08,DBLP:journals/pami/MoriBM05} have been intensively used in the domain of object recognition and image retrieval~\cite{DBLP:conf/icde/ZhangACT11,DBLP:journals/tip/XuSYSL17,DBLP:journals/pami/ShenXLYHS18} for decades as they are invariant to scaling, orientation and illumination changes. Consequently, a large amount of research efforts have been devoted to design effective features to facilitate ML tasks. Such a discipline was partially changed by the emergence of deep learning. In the past years, deep learning has transformed the world of artificial intelligence, due to the increasing processing power afforded by graphical processing units (GPUs), the enormous amount of available data, and the development of more advanced algorithms.  With well-annotated sufficient training data, the methodology can automatically learn the effective latent feature representation for the classification or prediction tasks. Hence, it can help replace the manual feature engineering process, which is non-trivial and labor-intensive.

In the area of automatic math problem solver, as reviewed in the previous sections, the only DL-based approach without feature engineering is DSN~\cite{DBLP:conf/emnlp/WangLS17}. It applies word embedding to vectorize the text information and encodes these vectors by GRU network for automatic feature extraction. The limitation is that a large amount of labeled training data is required to make the model effective. Before the appearance of DSN, most of the math problem solvers were designed with the availability of small-scale datasets. Thus, feature engineering plays an important role in these works to help achieve a high accuracy. In this section, we provide a comprehensive review on the features engineering process in the literature and show how they help bridge the gap between textual/visual information and the semantic/logical representation.

%In course of straightening out literature related to math word problems, we found that some features are used only in certain single work and some features are adopted cross different works, which are more principal and general. In the following, we will introduce features starting from where they appear across multiple works firstly, then to where they are adopted by single work.

\subsection{Preprocessing}

Before we review the feature space defined in various MWP solvers, we first present the preliminary background on the pre-processing steps that have been commonly adopted to facilitate the subsequent feature extraction.

\subsubsection{Syntactic Parsing}

Syntactic parsing focuses on organizing the lexical units and their semantic dependency in a tree structure, which serves as a useful resource for effective feature selection. Sorts of parsers have been developed, among which the Stanford parser works as the most comprehensive and widely-adopted one. It is a package consisting of different probabilistic natural language parsers. To be more specific, its neural-network parser~\cite{DBLP:conf/emnlp/ChenM14} is a transition-based dependency parser that uses high-order features to achieve high speed and good accuracy; the Compositional Vector Grammar parser ~\cite{DBLP:conf/acl/SocherBMN13} can be seen as factoring discrete and continuous parsing in one model; and the (English) Stanford Dependencies representation ~\cite{DBLP:conf/lrec/MarneffeMM06} is an automatic system to extract typed dependency parses from phrase structure parses, where a dependency parse represents dependencies between individual words and a phrase structure parse represents nesting of multi-word constituents. Besides Stanford parser, there exist other effective dependency parsers with their own traits. For example, ~\cite{Goldberg2010An} presents an easy-fist parsing algorithm that iteratively selects the best pair of neighbors in the tree structure to connect at each parsing step.

Those parsers account in WMP solvers. For instance, the neural-network parser~\cite{DBLP:conf/emnlp/ChenM14} is adopted in ~\cite{2017arXiv171209391R} for coreference resolution, which is another pre-processing step for MWP solvers. UnitDep~\cite{DBLP:conf/aaai/RoyR17} automatically generates features from a given math problem by analyzing its derived parser tree using the Compositional Vector Grammar parser~\cite{DBLP:conf/acl/SocherBMN13}. Additionally, the Stanford Dependencies representation~\cite{DBLP:conf/lrec/MarneffeMM06} has been applied in multitple solvers. We observed its occurrence in Formula~\cite{DBLP:conf/acl/MitraB16} and ARIS~\cite{DBLP:conf/emnlp/HosseiniHEK14} to extract attributes of entities (the subject, verb, object, preposition and temporal information), in KAZB~\cite{DBLP:conf/acl/KushmanZBA14} to generate part-of-speech tags, lematizations, and dependency parses to compute features, and in ALGES~\cite{DBLP:journals/tacl/Koncel-Kedziorski15} to obtain syntactic information used for grounding and feature computation. ExpressionTree~\cite{DBLP:conf/emnlp/RoyR15} is an exceptional case without using Stanford Parser. Instead, it uses the easy-fist parsing algorithm~\cite{Goldberg2010An} to detect the verb associated with each quantity.

\eat{~\cite{DBLP:journals/tacl/Koncel-Kedziorski15} introduced Qset, a compact representation of the properties of a quantity. Extracted from the dependency parse relations of a single sentence and for the future equation tree, a Qset is composed of quantity, entity, container, verb, and location. 
Likely, quantity schema~\cite{DBLP:conf/emnlp/RoyR15} is obtained on the basis of parse tree, providing all the information needed to solve most arithmetic problems. Features are further extracted from the components of the quantity schema, such as associated verb, subject, unit, related noun phrases and rate. Besides, ~\cite{DBLP:conf/aaai/Wang18,DBLP:conf/emnlp/HosseiniHEK14,DBLP:conf/acl/KushmanZBA14}  apply dependency parsers to generate features. Using tools like Stanford Parser,~\cite{DBLP:conf/aaai/Wang18}generated features automatically by analyzing its derived parser tree using, while [59] generated POS tags, lematizations, and dependency parses to compute features. Meanwhile, ARIS~\cite{DBLP:conf/emnlp/HosseiniHEK14} selected attributes of each entity from dependency parser. In addition to that,~\cite{2017arXiv171209391R} establishes feature functions based on dependency parse labels, which is used to extract related information from the problem combined with a small set of rules.}

\subsubsection{Coreference Resolution}

Co-reference resolution involves the identification and clustering of noun phrases mentions that refer to the same real-world entity. The MWP solvers use it as a pre-processing step to ensure the correct arithmetic operations or value update on the same entity. ~\cite{haghighi2009simple} is an early deterministic approach which is driven entirely by the syntactic and semantic compatibility  learned from a large, unlabeled corpus. It allows proper and nominal mentions to only corefer with antecedents that have the same head, but pronominal mentions to corefer with any antecedent. On top of~\cite{haghighi2009simple},  Raghunathan et al.~\cite{DBLP:conf/emnlp/RaghunathanLRCSJM10} proposed  an architecture based on tiers of deterministic coreference models. The tiers are processed from the highest to the lowest precision and the entity output of a tier is forwarded to the next tier for further processing.~\cite{DBLP:conf/conll/LeePCCSJ11} is another model that integrates a collection of deterministic co-reference resolution models. Targeting at exploring rich feature space,~\cite{Bengtson:2008:UVF:1613715.1613756} proposed a simple classification model for coreference resolution with a well-designed set of features. NECo is proposed in ~\cite{DBLP:conf/emnlp/HajishirziZWZ13} and capable of solving both named entity linking and co-reference resolution jointly.  %Likely in a rich feature space,~\cite{Lee:2013:DCR:2576217.2576221}  performed entity-centric co-reference using solely simple, deterministic rules. As a principled but simple framework for co-reference resolution, the Latent Left Linking model ~\cite{ChangSaRo13} is optimized by a fast stochastic gradient-based technique. 

As to applying coreference resolvers in MWP sovers, the Illinois Coreference Resolver~\cite{Bengtson:2008:UVF:1613715.1613756}~\cite{ChangSaRo13} is used in ~\cite{DBLP:journals/tacl/RoyVR15} to identify pronoun referents and facilitate semantic labeling. Alternatively,  a rule function $\mathtt{Coref(A,B)}$, which is true when A and B represent the same entity, is derived in~\cite{2017arXiv171209391R} as a component of the declarative rules to determine the math operators. Given a pair of sentences, each containing a quantity, ZDC~\cite{DBLP:conf/emnlp/ZhouDC15} takes into account the existence of coreference relationship between these two sentences for feature exploitation. Meanwhile, ARIS~\cite{DBLP:conf/emnlp/HosseiniHEK14} adopts the ~\cite{DBLP:conf/emnlp/RaghunathanLRCSJM10} for coreference resolution and uses the predicted coreference relationships to replace pronouns with their coreferenent links.

\subsection{Common Features}
There have been various types of features proposed in the past literature. We separate them into \textit{common features} and \textit{unique features}, according to the number of solvers that have adopted a particular type of feature. The unique features were proposed once and not reused in another work, implying that their effect could be limited. The \textit{common features} are considered to be more general and effective, and they are the focus of this survey.

In Table~\ref{tbl:common-feature}, we categorize the \textit{common features} according to their syntactic sources for feature extraction, such as quantities, questions, verbs, etc. For each type of proposed feature, we identify its related MWP solvers, and provide necessary examples to explain features that are not straightforward to figure out.

\begin{table*}[!htbp] 
\centering
\caption{Common Features.} 
\label{tbl:common-feature}
\begin{tabular}{|p{2cm}|p{4.5cm}|p{2.5cm}|p{7.5cm}|}
\hline
Feature Type &  Description & Used In & Remark \\ \hline
\multirow{3}{2cm}{Quantity-related Features} & Whether the quantity refers to a rate & \cite{DBLP:conf/emnlp/RoyR15}~\cite{DBLP:conf/aaai/RoyR17}~\cite{DBLP:conf/aaai/Wang18}~\cite{DBLP:conf/emnlp/ZhouDC15}~\cite{DBLP:conf/emnlp/UpadhyayCCY16} & For ``each ride cost 5 tickets'', the quantity ``5'' is a rate\\ \cline{2-4}
%& Position of a number w.r.t a comparative word & ~\cite{DBLP:conf/emnlp/ZhouDC15}~\cite{DBLP:conf/emnlp/UpadhyayCCY16} & \\ \cline{2-4}
& Is between 0 and 1 & \cite{DBLP:conf/emnlp/ZhouDC15} ~\cite{DBLP:conf/emnlp/UpadhyayCCY16}&\\ \cline{2-4}
& Is equal to one or two & \cite{DBLP:conf/acl/KushmanZBA14}~\cite{DBLP:conf/emnlp/ZhouDC15}~\cite{DBLP:conf/emnlp/UpadhyayCCY16}~\cite{DBLP:conf/emnlp/HuangSYL17} & \\ \cline{2-4}
\hline

\multirow{4}{2cm}{Context-related Features} & Word lemma  & \cite{DBLP:conf/acl/KushmanZBA14}~\cite{DBLP:conf/emnlp/ZhouDC15}~\cite{DBLP:conf/emnlp/UpadhyayCCY16} & For ``Connie has 41.0 red markers.'', the word lemmas around the quantity ``41.0'' are \{Connie, have, red, marker\}. \\ \cline{2-4}
&POS tags & \cite{DBLP:conf/aaai/RoyR17}~\cite{DBLP:conf/acl/KushmanZBA14}~\cite{DBLP:conf/emnlp/ZhouDC15}~\cite{DBLP:conf/emnlp/UpadhyayCCY16}~\cite{DBLP:journals/tacl/RoyVR15} & For ``A chef needs to cook 16.0 potatoes.'', the POS tags within a window of size 2 centered at the quantity ``16.0'' are \{TO, VB, NNS\}.\\ \cline{2-4}
& Dependence type & \cite{DBLP:conf/acl/KushmanZBA14}~\cite{DBLP:conf/emnlp/ZhouDC15}~\cite{DBLP:conf/emnlp/UpadhyayCCY16} & For ``Ned bought 14.0 boxes of chocolates candy.'', we can detect multiple dependencies within the window of size 2 around the ``14.0'': (boxes, 14.0) $\rightarrow$ (num), (boxes, of)$\rightarrow$ (prep), (bought, Ned) $\rightarrow$ (nsubj). The dependence root is ``bought''. \\ \cline{2-4}
& Comparative adverbs & \cite{DBLP:conf/emnlp/RoyR15}~\cite{DBLP:conf/aaai/RoyR17}~\cite{DBLP:conf/aaai/Wang18} &For ``If she drank 25 of them and then bought 30 more.'', ``more'' is a comparative term in the window of quantity ``30''.
\\ \hline

\multirow{7}{2cm}{Quantity-pair Features} & 
Whether both quantities have the same unit  & \cite{DBLP:conf/emnlp/RoyR15}~\cite{DBLP:conf/aaai/RoyR17}~\cite{DBLP:conf/aaai/Wang18}~\cite{DBLP:conf/emnlp/ZhouDC15} & For ``Student tickets cost 4 dollars and general admission tickets cost 6 dollars'', quantities ``4'' and ``6'' have the same unit.\\ \cline{2-4}
&If one quantity is related to a rate and the other is associated with a unit that is part of the rate & \cite{DBLP:conf/emnlp/RoyR15}~\cite{DBLP:journals/tacl/Koncel-Kedziorski15}~\cite{DBLP:conf/aaai/RoyR17}~\cite{DBLP:conf/aaai/Wang18} & For ``each box has 9 pieces'' and ``Paul bought 6 boxes of chocolate candy'',  ``9'' is related to a rate ( i.e., pieces/box) and ``6'' is associated to the unit ``box''.\\ \cline{2-4}
&Numeric relation of two quantities & \cite{DBLP:conf/emnlp/ZhouDC15}~\cite{DBLP:conf/emnlp/UpadhyayCCY16} & For each quantity, the nouns around it are extracted and sorted by the distance in the dependency tree. Then, a scoring function is defined on the two sorted lists to measure the numeric relation. \\ \cline{2-4}
&Context similarity between two quantities & \cite{DBLP:conf/emnlp/ZhouDC15} ~\cite{DBLP:conf/emnlp/UpadhyayCCY16}& The context is represented by the set of words around the quantity. \\ 
\cline{2-4}
&Dependency path between two quantities. & \cite{DBLP:conf/acl/KushmanZBA14}~\cite{DBLP:conf/emnlp/ZhouDC15} & For ``2 footballs and 3 soccer balls cost 220 dollars'', the dependency path for the quantity pair $(2,3)$ is \emph{num}(footballs,2) -- \emph{conj}(footballs, balls) -- \emph{num}(balls, 3). \\ \cline{2-4}
& Whether both quantities appear in the same sentence & \cite{DBLP:conf/acl/KushmanZBA14}~\cite{DBLP:conf/emnlp/ZhouDC15} &  \\ \cline{2-4}
& Whether the value of the first quantity is greater than the other  & \cite{DBLP:conf/emnlp/RoyR15}~\cite{DBLP:conf/aaai/RoyR17}~\cite{DBLP:conf/aaai/Wang18}~\cite{DBLP:conf/acl/KushmanZBA14} \cite{DBLP:conf/emnlp/ZhouDC15}~\cite{DBLP:conf/emnlp/UpadhyayCCY16} &\\ 
\hline

{\multirow{7}{2cm}{Question-related Features}} & 
Whether the unit or related noun phrase of a quantity appears in the question & \cite{DBLP:journals/tacl/RoyVR15}~\cite{DBLP:conf/emnlp/RoyR15}~\cite{DBLP:journals/tacl/Koncel-Kedziorski15}~\cite{DBLP:conf/aaai/RoyR17}~\cite{DBLP:conf/aaai/Wang18} ~\cite{DBLP:conf/acl/KushmanZBA14} &  \\ \cline{2-4}
&Whether the unit or related noun phrase of a quantity has the highest number of match tokens with the question text & \cite{DBLP:conf/emnlp/RoyR15}~\cite{DBLP:conf/aaai/RoyR17}~\cite{DBLP:conf/aaai/Wang18} & For the question ``How many apples are left in the box?'' and a quantity $77$ that appears in ``77 apples in a box'', there are two matching tokens (''apples'' and ``box'').\\ \cline{2-4}
& Number of quantities which happen to have the maximum number of matching tokens with the question & \cite{DBLP:conf/emnlp/RoyR15}~\cite{DBLP:conf/aaai/RoyR17}~\cite{DBLP:conf/aaai/Wang18} & For ``Rose have 9 apples and 12 erasers. ... 3 friends. How many apples dose each friend get?'', the number of matching tokens for quantities $9$, $12$ and $3$ is $1$, $0$ and $1$. Hence, there are two quantities with the maximum matching token number.\\ \cline{2-4}
%&Numberic relation between numbers and the question sentence using the surrounding nouns & \cite{DBLP:conf/emnlp/ZhouDC15} ~\cite{DBLP:conf/emnlp/UpadhyayCCY16} & For the question ``How many work-stations accommodate 2 students?'', nouns list ``students, time, chemistry, lab'' is extracted according to dependency path for the ``38''. the relation evaluated by the reciprocal of index of ``students'' in the noun list is equal to $1/1$   \\ \cline{2-4}
&Whether any component of the rate is present in the question& \cite{DBLP:conf/emnlp/RoyR15}~\cite{DBLP:conf/aaai/RoyR17}~\cite{DBLP:conf/aaai/Wang18} & Given a question ``How many blocks does George have?'' and a quantity $6$ associated with rate ``blocks/box'', the feature indicator is set to $1$ since block appears in the question. \\ \cline{2-4}
&Whether the question contains terms like ``each'' or ``per''  & \cite{DBLP:conf/emnlp/RoyR15}~\cite{DBLP:conf/aaai/RoyR17}~\cite{DBLP:conf/aaai/Wang18}& \\ \cline{2-4}
&Whether the question contains comparison-related terms like ``more'' or ``less''  & \cite{DBLP:conf/emnlp/RoyR15}~\cite{DBLP:conf/aaai/RoyR17}~\cite{DBLP:conf/aaai/Wang18} & \\ 
&Whether the question contains terms like ``how many'' &\cite{DBLP:conf/acl/KushmanZBA14}~\cite{DBLP:conf/emnlp/ZhouDC15}~\cite{DBLP:conf/emnlp/UpadhyayCCY16}~\cite{DBLP:conf/emnlp/HuangSYL17} & It implies that the solution is positive. \\ \cline{2-4}
\hline

{\multirow{4}{2cm}{Verb-related Features}} 
& Dependent verb of a quantity & \cite{DBLP:conf/emnlp/RoyR15}~\cite{DBLP:journals/tacl/Koncel-Kedziorski15}~\cite{DBLP:conf/aaai/RoyR17}~\cite{DBLP:conf/aaai/Wang18} & the verb closest to the quantity in the dependency tree\\ \cline{2-4}
& Distance vector between the dependent verb and a small collection of pre-defined verbs that are useful for arithmetic operator classification & \cite{DBLP:conf/emnlp/HosseiniHEK14} ~\cite{Liang16} ~\cite{DBLP:journals/tacl/Koncel-Kedziorski15} & \\ \cline{2-4}
& Whether two quantities have the same dependent verbs & \cite{DBLP:conf/emnlp/RoyR15}~\cite{DBLP:conf/aaai/RoyR17}~\cite{DBLP:conf/aaai/Wang18}& For ``In the first round she scored 40 points and in the second round she scored 50 points'', the quantities ``40'' and ``50'' both have the same verb ``scored''. Note that ``scored'' appeared twice in the sentence.\\ \cline{2-4}
& Whether both dependent verbs refer to the same verb mention & \cite{DBLP:conf/emnlp/RoyR15}~\cite{DBLP:conf/aaai/RoyR17}~\cite{DBLP:conf/aaai/Wang18} & For ``She baked 4 cupcakes and 29 cookies.'', the quantities ``4'' and ``29'' both shared the verb ``baked''. Note that ``baked'' appeared only once in the sentence.\\  
\hline

{\multirow{4}{2cm}{Global Features}} & 
Number of quantities mentioned in text& \cite{DBLP:conf/emnlp/RoyR15}~\cite{DBLP:conf/aaai/RoyR17}~\cite{DBLP:conf/aaai/Wang18} & \\ \cline{2-4}
& Unigrams and bigrams of sentences in the problem text &\cite{DBLP:journals/tacl/RoyVR15}~\cite{DBLP:conf/acl/KushmanZBA14} & \\ \cline{2-4}
%&Various conjunctions of the all features & ~\cite{DBLP:journals/tacl/RoyVR15}~\cite{DBLP:conf/emnlp/RoyR15}~\cite{DBLP:conf/aaai/RoyR17}~\cite{DBLP:conf/aaai/Wang18}&\\ 
\hline
\end{tabular}
\end{table*}

%Based on our observation, features adopted across different works are often more principal and general than by single work. Similar to section 5.1, we introduce these features by seperating them into different categories, such as single-quantity, quantity-pair, related to questions, related to verb, and others.

\subsubsection{Quantity-related Features}
The basic units in an arithmetic expression or an equation set consist of quantities, unknown variables and operators. Hence, a natural idea is to extract quantity-related features to help identify the relevant operands and their associated operators. As shown in Table~\ref{tbl:common-feature}, a binary indicator to determine whether a quantity refers to a rate is adopted in many solvers~\cite{DBLP:conf/emnlp/RoyR15}~\cite{DBLP:conf/aaai/RoyR17}~\cite{DBLP:conf/aaai/Wang18}~\cite{DBLP:conf/emnlp/ZhouDC15}~\cite{DBLP:conf/emnlp/UpadhyayCCY16}. It signals a strong connection between the quantity and operators of $\{\times,\div\}$. The value of the quantity is also useful for operator classifier or quantity relevance classifier. For instance, a quantity whose value is a real number between $[0,1]$ is likely to be associated with multiplication or division operators~\cite{DBLP:conf/emnlp/ZhouDC15},~\cite{DBLP:conf/emnlp/UpadhyayCCY16}. It is also observed that quantities in the text format of ``one'' or ``two'' are unlikely to be relevant with the solution~\cite{DBLP:conf/acl/KushmanZBA14}~\cite{DBLP:conf/emnlp/ZhouDC15},~\cite{DBLP:conf/emnlp/UpadhyayCCY16},~\cite{DBLP:conf/emnlp/HuangSYL17}. Examples include ``if \textit{one} airplane averages 400 miles per hour,...'' and ``the difference between \textit{two} numbers is 36''.

%In general, the quantities always come with their units following by them in text, for instance, in ``She scored 50 points'', ``point'' is the unit of ``50''. There are also cases that more than one unit that could be extracted in the dependency path of some quantities, as illustrated in the follows. Two different units may imply that the quantity is possibly a rate, like ``ride'' and ``tickets'' in ``each ride cost 5 tickets''. Whether the quantity is a rate is used as a feature in ~\cite{DBLP:conf/emnlp/RoyR15},~\cite{DBLP:conf/aaai/RoyR17},~\cite{DBLP:conf/aaai/Wang18},~\cite{DBLP:conf/emnlp/ZhouDC15},~\cite{DBLP:conf/emnlp/UpadhyayCCY16} to work for multiplication and division. 

%It's worth noting that the words around the quantities, position of the quantities, and the syntactic structure associated with quantities are often adequately informative, thus unigrams, bigrams and part of speech tags within a window around the quantity are added as features in many algorithms like ~\cite{DBLP:conf/acl/KushmanZBA14},~\cite{DBLP:conf/emnlp/ZhouDC15},~\cite{DBLP:conf/emnlp/RoyR15},~\cite{DBLP:conf/aaai/RoyR17},~\cite{DBLP:conf/emnlp/UpadhyayCCY16}.

\subsubsection{Context-related Features}
The information embedded in the text window centered at a particular quantity can also provide important clues for solving math word problems. To differentiate two quantities  both in the numeric format, we can leverage the word lemmas, part of speech (POS) tags and dependence types within the window as the features. In this manner, quantities associated with the same operators would to likely to share similar context information. A trivial trick used in~\cite{DBLP:conf/emnlp/RoyR15}~\cite{DBLP:conf/aaai/RoyR17}~\cite{DBLP:conf/aaai/Wang18} is to examine whether there exists comparative adverbs. For example, terms ``more'', ``less'' and ``than'' indicate operators of $\{+,-\}$.

\subsubsection{Quantity-pair Features}
The relationship between two quantities is helpful to determine their associated operator. A straightforward example is that if two quantities are associated with the same unit, they can be applied with addition and subtraction~\cite{DBLP:conf/emnlp/RoyR15}~\cite{DBLP:conf/aaai/RoyR17}~\cite{DBLP:conf/aaai/Wang18}~\cite{DBLP:conf/emnlp/ZhouDC15}. If one quantity is related to a rate and the other is associated with a unit that is part of the rate, their operator is likely to be multiplication or division~\cite{DBLP:conf/emnlp/RoyR15}~\cite{DBLP:journals/tacl/Koncel-Kedziorski15}~\cite{DBLP:conf/aaai/RoyR17}~\cite{DBLP:conf/aaai/Wang18}. 

Numeric relation and context similarity are two types of quantity-pair features proposed in ~\cite{DBLP:conf/emnlp/ZhouDC15}~\cite{DBLP:conf/emnlp/UpadhyayCCY16}. The former obtains two sets of nouns located within the same sentence as the two quantities and sorts them by the distance in the dependency tree. Then, a scoring function is defined to measure the similarity between these two sorted noun lists. Higher similarity implies that the two quantities are more likely to be connected by addition or subtraction operators. The latter extracts features for equation template classifier.  It is observed that the contextual information between two numbers is similar, they are likely to be located within in a template with  symmetric number slots. For example, given a template $\mathtt{n_1\times u_1 + n_2\times u_2}$, ``$n_1$'' and ``$n_2$'' are symmetric. The context similarity is measured by the Jaccard similarity on two sets of words among the context windows. Given a problem text ``A plum costs 2 dollars and a peach costs 1 dollars'', ``2'' and ``1'' are two quantities with similar context.

Two types of quantity-pair features were both adopted in the template-based solutions to equation set problems~\cite{DBLP:conf/acl/KushmanZBA14}~\cite{DBLP:conf/emnlp/ZhouDC15}. The first type is the dependency path between a pair of quantities. Their similarity may be helpful to determine the corresponding positions (or number slots) in the equation template. For example, given a sentence ``2 footballs and 3 soccer balls cost 220 dollars'', the dependency paths between two quantity pairs $(2,220)$ and $(3,220)$ are identical, implying that $2$ and $3$ refer to similar types of number slots in the template. The other feature is whether two quantities appear in the same sentence. If so, they are likely to appear in the same equation of the template. Finally, a popular quantity-pair feature used in  ~\cite{DBLP:conf/emnlp/RoyR15}~\cite{DBLP:conf/aaai/RoyR17}~\cite{DBLP:conf/aaai/Wang18}~\cite{DBLP:conf/acl/KushmanZBA14} ~\cite{DBLP:conf/emnlp/ZhouDC15}~\cite{DBLP:conf/emnlp/UpadhyayCCY16} examines whether the value of one quantity is greater than the other, which is helpful to determine the correct operands for subtraction operator.

\subsubsection{Question-related Features}
Distinguishing features can also be derived from questions. It is straightforward to figure out that the unknown variable can be inferred from the question and if a quantity whose unit appears in the question, this quantity is likely to be relevant. The remain question-related features presented in Table~\ref{tbl:common-feature} were proposed by Roy et al.~\cite{DBLP:conf/emnlp/RoyR15,DBLP:conf/aaai/RoyR17} and followed by MathDQN~\cite{DBLP:conf/aaai/Wang18}. Their feature design leverages the number of matching tokens between the related noun phrase of a quantity and the question text. The quantities with the highest number of matching tokens are considered as useful clues. They also check whether the question contains rate indicators such as ``each'' and ``per'', or comparison indicators such as ``more'' or ``less''. The former is related to $\{\times,\div\}$ and the latter is related to $\{+,-\}$. Moreover, if the question text contains ``how many'', it implies that the solution is a positive number.

\subsubsection{Verb-related Features}
Verbs are important indicators for correct operator determination. For example, ``lose'' is a verb indicating quantity loss for an entity and related to the subtraction operator. Given a quantity, we call the verb closest to it in the dependency tree as its \textit{dependent verb}. ~\cite{DBLP:conf/emnlp/RoyR15}~\cite{DBLP:journals/tacl/Koncel-Kedziorski15}~\cite{DBLP:conf/aaai/RoyR17}~\cite{DBLP:conf/aaai/Wang18} directly use dependent verb as one of the features. Another widely-adopted verb-related feature is a vector capturing the distance between the dependent verb and a small pre-defined collection of verbs that are found to be useful in categorizing arithmetic operations. Again, the remaining features come from the works~\cite{DBLP:conf/emnlp/RoyR15,DBLP:conf/aaai/RoyR17,DBLP:conf/aaai/Wang18}. The features indicate whether two quantities have the same dependent verbs or whether their dependent verbs refer to the same verb mention. As we can see from the examples in Table~\ref{tbl:common-feature}, the difference between these two types of features is the occurrence number of the dependent verb in the sentence. 
  
% used the dependent verbs of quantities as the features directly. And the similarity is computed in ~\cite{DBLP:conf/emnlp/HosseiniHEK14} ~\cite{Liang16} ~\cite{DBLP:journals/tacl/Koncel-Kedziorski15} between the verb and a list of seed verbs which are found to be useful in categorizing arithmetic operations. When the two quantities have same verb, other information may help to determine the operator between them. For example, for ``In the first round she scored 40 points and in the second round she scored 50 points'',  the question asking ``how many more points is the first round than the second round'' lead to different arithmetic operator between ``40'' and ``50'' from asking ``total''. Thereby, whether the pair quantities have the same dependent verbs or the same verb mention are added as features in ~\cite{DBLP:conf/emnlp/RoyR15},~\cite{DBLP:conf/aaai/RoyR17},~\cite{DBLP:conf/aaai/Wang18}.

\subsubsection{Global Features}
There are certain types of global features in the document-level proposed by existing solvers. \cite{DBLP:conf/emnlp/RoyR15,DBLP:conf/aaai/RoyR17,DBLP:conf/aaai/Wang18} use the number of quantities in the problem text as part of feature space. Unigrams and bigrams are also applied in~\cite{DBLP:journals/tacl/RoyVR15}~\cite{DBLP:conf/acl/KushmanZBA14}. They may play certain effect in determining the quantities and their order. Note that the unigrams and bigrams are defined in the word level rather than the character level.

\section{Geometric Word Problem (GWP) Solver}\label{sec:geometric}
Geometry solvers have been studied for a long history. Visual diagram understanding is a sub-domain that has attracted significant attention. As an early work for understanding line drawings, ~\cite{ DBLP:journals/cvgip/LinSMS85} presented an efficient characteristic pattern detection method  by scanning the distribution of black pixels and generating feature points graph. A structure mapping engine named \textit{GeoRep} was proposed in~\cite{DBLP:conf/aaai/FergusonF00} to generate qualitative spatial descriptions from line diagrams. After that, the visual elements can be formulated through a two-level representation architecture. This work was also applied to the repetition and symmetry detection model in MAGI~\cite{ Ferguson98tellingjuxtapositions:}. Inspired by human cognitive process of reading juxtaposition diagrams, MAGI detects repetition by aligning visual and conceptual relational structure to analyze repetition-based diagrams.

The problem of rectangle and parallelogram detection in diagram understanding has also received a considerable amount of interest. The proposed techniques fall into two main categories, either based on primitive or Hough transform~\cite{DBLP:journals/cacm/DudaH72}. The primitive-based methods combine line segments or curves to form possible edges of a quadrangle. For examples, Lin and Nevatia ~\cite{ DBLP:journals/cviu/LinN98} proposed the approach of parallelogram detection from a single aerial image by linear feature extraction and formation of hypothesis following certain geometric constraints. Similarly, Lagunovsky and Ablameyko~\cite{DBLP:journals/prl/LagunovskyA99} studied the problem of rectangular detection based on line primitives. As to the Hough transform based techniques, \cite{DBLP:journals/tmi/ZhuCMP03} presented an approach for automatic rectangular particle detection in cryo-electron microscopy through Hough transform, but this method can only work well when all rectangles have the same dimensions and the dimensions must be aware in advance. Jung et.al.~\cite{DBLP:conf/sibgrapi/JungS04} proposed a window Hough transform algorithm to tackle the problem of rectangle detection with varying dimensions and orientations.

%%%%%%%%%%%%%%%%%%%%
Geometry theorem proving (GTP)~\cite{Gelernter:1995:RGP:216408.216418,Fleuriot2001} was initially viewed as an artificial intelligence problem that was expected to be easily tackled by machines. The difficulties of solving GTP problems lie in the visual reasoning in geometry and the generation of elegant and concise geometry proofs. Moreover, completing the proof requires the ingenuity and insights to the problem. The first automated GTP was developed by Gelernter ~\cite{Gelernter:1995:RGP:216408.216418}, which used the diagram as pruning heuristic. The system rejects geometry goals that fail to satisfy the diagram. Whereas the limitation of the method is the true sub-goal may be pruned erroneously due to the insufficient precise arithmetic applied to the diagram. Fleuriot et. al.~\cite{ Fleuriot2001} studied  Newton's geometric reasoning procedures in his work Principia and presented theorem prover Isabelle to formalize and generate the style of reasoning performed by Newton. By combining the existing geometry theorem proving techniques and the concepts of Nonstandard Analysis, the prover Isabelle can produce proofs of lemmas and theorem in Principia. Readers can refer to the book chapter~\cite{Fleuriot2001} for the survey of early development of GTP.

\begin{figure}[h]
\centering
\includegraphics[width=68mm, keepaspectratio]{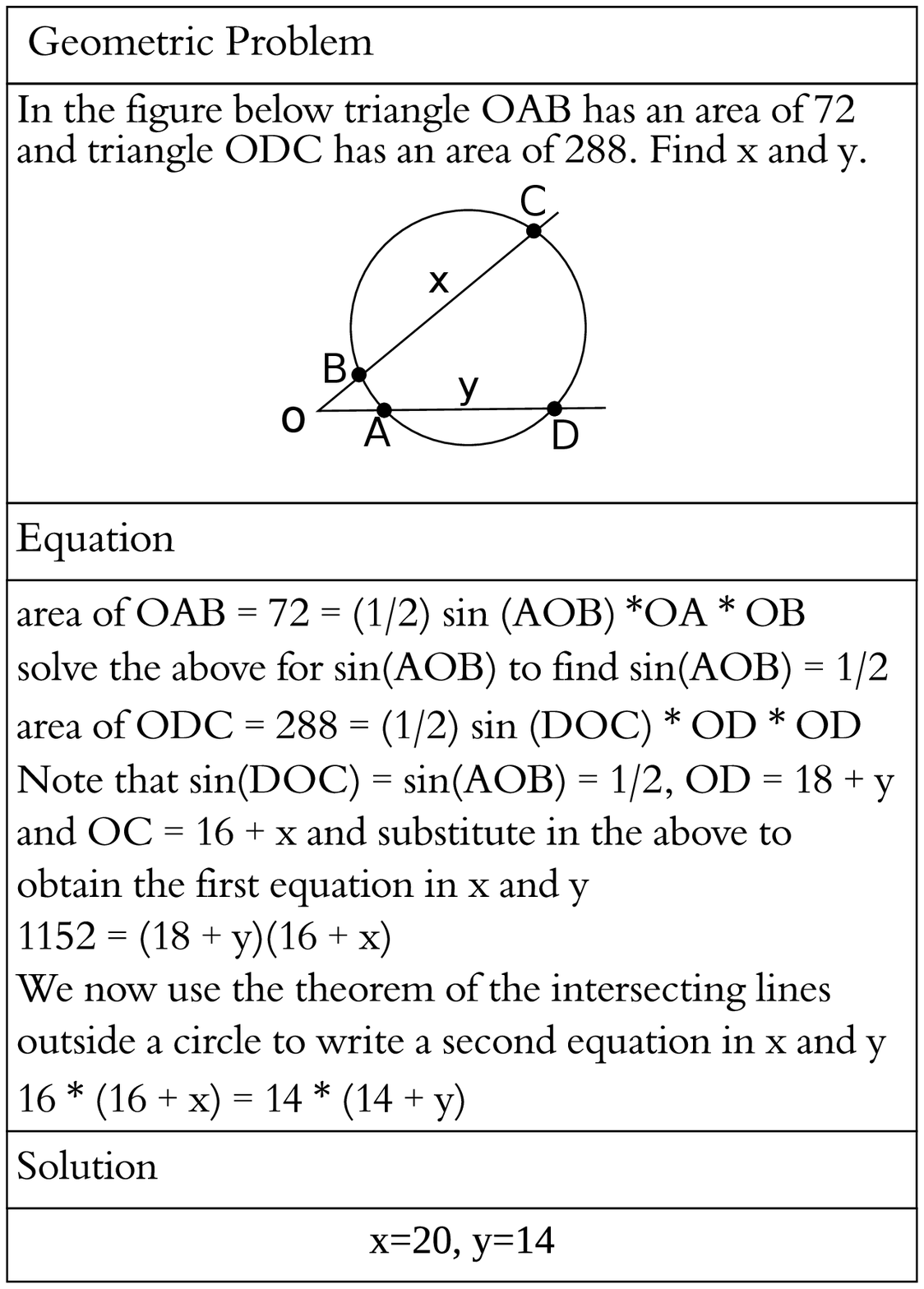} 
\caption{An example of geometric problem.}
\label{fig:example-geometric}
\end{figure}

In this survey, we are more interested to examine the math problems that are required to consider visual diagram and textual mentions simultaneously. As illustrated in Figure~\ref{fig:example-geometric},  a typical geometry word problem contains text descriptions or attribute values of geometric objects. The visual diagram may contain essential information that are absent from the text. For instance, points $O$, $B$ and $C$ are located on the same line segment and there is a circle passing points $A,B,C$ and $D$. To well solve geometry word problems, three main challenges need to be tackled: 1) diagram parsing requires the detection of visual mentions, geometric characteristics, the spatial information and the co-reference with text; 2) deriving visual semantics which refer to the textual information related to visual analogue involves the semantic and syntactic interpretation to the text; and 3) the inherent ambiguities lie in the task of mapping visual mentions in the diagram to the concepts in real world.

%The problem is challenging because it needs to be mapped into a logical representation that is compatible with both the problem text and the accompanying diagram. Common strategies to solve geometry word problems constitute three key components, including diagram understanding to capture visual clues, text parsing to capture semantic information, and deductive reasoning via a knowledge base with geometry axioms and theorems.

\subsection{Text-Aligned Diagram Understanding}
%\cite{ DBLP:conf/ieaaie/Bulko88},~\cite{ DBLP:conf/aaai/NovakB90},~\cite{ DBLP:journals/air/Srihari94},~\cite{ DBLP:conf/acl/WatanabeN98} presented a framework of integrating textual and visual mentions for diagram understanding.

The very early computer program, BEATRIX~\cite{ DBLP:conf/ieaaie/Bulko88,DBLP:conf/aaai/NovakB90}, parses the English text and diagram components of the elementary physics problems together by establishing the coreference between the text and diagram. Watanabe et al. proposed a framework to combine layout information and natural language to analyze the pictorial book of flora diagrams~\cite{DBLP:conf/acl/WatanabeN98}. An overview of the research on integration of visual and linguistic information was provided in the survey paper by Srihar~\cite{DBLP:journals/air/Srihari94}. However, these early approaches rely on written rules or manual regulations, i.e., the visual elements needed to be recognized with human intervention and their performances were usually dependent on specified diagrams.

%~\cite{DBLP:conf/lpar/ItzhakyGIS13} combines deductive, numeric, and inductive reasoning to solve geometric problems. First, the partial program is generated from logical constraints by deductive reasoning and then evaluated by numerical methods to produce partial solutions to the geometry problem. Next, new constraints are added by inductive synthesis and applied to the next round until the complete program is generated. Since it is an interactive system, cooperation between humans and machines is required.

G-ALINGER~\cite{DBLP:conf/aaai/SeoHFE14} is an algorithmic work that addresses the geometry understanding and text understanding simultaneously. To detect primitives from a geometric diagram, Hough transform~\cite{Shapiro01} is first applied to initialize lines and circles segments. An objective function that incorporates pixel coverage, visual coherence and textual-visual alignment. The function is sub-modular and a greedy algorithm is designed to pick the primitive with the maximum gain in each iteration. The algorithm stops when no positive gain can be obtained according to the objective function. In~\cite{DBLP:conf/eccv/KembhaviSKSHF16}, the problem of visual understanding is addressed in the context of science diagrams. The objective is to identify the graphic representation for the visual entities and their relations such as temporal transitions, phase transformations and inter object dependencies. An LSTM-based network is proposed for syntactic parsing of diagrams and learns the graphic structure.

\subsection{GWP Solvers}

GEOS~\cite{DBLP:conf/emnlp/SeoHFEM15} can be considered as the first work to tackle a complete geometric word problem as shown in Figure~\ref{fig:example-geometric}. The method consists of two main steps: 1) parsing text and diagram respectively by generating a piece of logical expression to represent the key information of the text and diagram as well as the confidence scores, and 2) addressing the optimization problem by aligning the satisfiability of the derived logical expression in a numerical method that requires manually defining indicator function for each predicate. It is noticeable that G-ALINGER is applied in GEOS~\cite{DBLP:conf/aaai/SeoHFE14} for primitive detection. Despite the superiority of automated solving process, the performance of the system would be undermined if the answer choices are unavailable in a geometry problem and the deductive reasoning based on geometric axiom is not used in this method. A subsequent improver of GEOS is presented in~\cite{DBLP:conf/emnlp/SachanDX17}. It harvests an axiomatic knowledge from $20$ publicly available math textbooks and builds a more powerful reasoning engine that leverages the structured axiomatic knowledge for logical inference.

GeoShader~\cite{DBLP:conf/aied/AlvinGMM17}, as the first tool to automatically handle geometry problem with shaded area, presents an interesting reasoning technique based on analysis hypergraph. The nodes in the graph represent intermediate facts extracted from the diagram and the directed edges indicate the relationship of deductibility between two facts. The calculation of the shaded area is represented as the target node in the graph and the problem is formulated as finding a path in the hypergraph that can reach the target node.

%\subsubsection{Structure-Based Features}

%Structural features tend to encode the syntactic cues such as the similarity between the textual mentions ~\cite{DBLP:conf/emnlp/SachanDX17}, dependency tree edge labels ~\cite{DBLP:conf/emnlp/SeoHFEM15}, word distances ~\cite{DBLP:conf/emnlp/SeoHFEM15}, or the typographical contents ~\cite{DBLP:conf/emnlp/SachanDX17}. The features in a geometry proof problems synthesizing framework ~\cite{DBLP:conf/aaai/AlvinGMM14} are defined as the size and type of the goal set as well as quantitative features of the proof such as the depth, width or the number of hyperedges in a proof. 

%\subsubsection{Content-Based Features}

%Content based features capture semantic cues such as the similarity between the sentences in the text  ~\cite{DBLP:conf/emnlp/SachanDX17}and the elements related to geometry language. \cite{DBLP:conf/aaai/SeoHFE14} defined several visual elements to be the combination of primitives that denote line or circle segment. Such expressions were further formalized in the work of ~\cite{DBLP:conf/emnlp/SeoHFEM15}, which defined geometry entities to be constants, variables, predicates, functions and etc. Based on the previous work, ~\cite{DBLP:conf/emnlp/SachanDX17} added more logical expressions to their parsing model and obtained $19$ entities and $115$ functions and predicates compared with $13$ entity types and $94$ functions and predicates used in GEOS.

\section{Miscellaneous Math Tasks}
\label{sec:misc}

\subsection{Word Problems in Related Domains}
Apart from geometric problems, there are also assorted variants of math problems that AI system focuses on. Aristo~\cite{DBLP:conf/aaai/ClarkEKSTTK16} is able to solve non-diagram multiple-choice questions through five parallel solvers, one for pure text, two for statistic and two for inference. Finally, the combiner of Aristo outputs a comprehensive score of each option based on scores from the five solvers. A similar work on multiple-choice questions is \cite{DBLP:conf/ijcai/ChengZWCQ16}, which takes Wikipedia as a knowledge base. After ranking and filtering relevant pages retrieved from Wikipedia, it presents a new scoring function to pick the best answer from the candidates. Another vairant is targeted at solving IQ test and a noticeable number of computer models  have been proposed in~\cite{DBLP:conf/ijcai/Hernandez-Orallo17,DBLP:conf/emnlp/WangTGZBL16}. Taking~\cite{DBLP:conf/emnlp/WangTGZBL16} for example, it proposed a framework for solving verbal IQ questions, which classifies questions into several categories and each group of questions are solved by a specific solver respectively. Furthermore, logic puzzles are addressed in~\cite{Lev:2004:SLP:1628275.1628277} by transforming robust natural language to precise semantics. For other forms of math problems, \cite{DBLP:conf/ijcai/DriesKDBR17} solves probability problems automatically by a two-step approach, namely first formulating questions in a declarative language and then computing the answer through a solver implemented in ProbLog~\cite{DeRaedt:2007:PPP:1625275.1625673}. And algebraic word problems are solved by generating answer rationales written in natural language in~\cite{DBLP:conf/acl/LingYDB17} through a sequence-to-sequence model.

\subsection{Math Problem Solver in Other Languages}
Solving math word problems in other languages also attracts research attention. Yu et al. addressed the equation set problem solver in Chinese~\cite{7446146,7839530}. Syntax-semantics (S$^2$) model was proposed to extract quantity relations from a given problem.  Each model contains information about keyword structure, pattern of POS and quantity relation. Compared with solutions that map sentences into pre-defined templates, the works show that they can use fewer number of models in the semantic parsing.  The experiments were conducted on a very small-scale dataset with $104$ problems. Recently, there has been the first attempt to solve Arabic arithmetic word problems~\cite{DBLP:conf/acling/SiyamSAS17}. Its test dataset was collected by translating the \textbf{AI2} dataset~\cite{DBLP:conf/emnlp/HosseiniHEK14} from English to Arabic. The proposed techniques also rely on the verb categorization, similar to those proposed in~\cite{DBLP:conf/emnlp/HosseiniHEK14}, except that customization for the Arabic language needs to be made for the tasks of syntactic parser and named entity recognition. To conclude, the math word problem solvers in other languages than English are still at a very early stage. The datasets used are neither large-scale nor challenging and the proposed techniques are obsolete. This research area has great room for improvement and calls for more efforts to be involved.

\subsection{Math Problem Generator}
We also review automatic math word problem generators that can efficiently produce a large, diverse and configurable corpus of question-answer database. The topics covered in this survey include algebra word problems with basic operators $\{+,-,\times,\div\}$ and geometry problems.

In~\cite{DBLP:conf/ijcai/WangS16}, Wang et al. leveraged the concept of \textit{expression tree} to generate a math word problem. The tree structure can provide the skeleton of the story, and meanwhile allow the story to be constructed recursively from the sub-stories. Each sub-story can be seen as a text template with value slots to be filled. These sub-stories will be concatenated into an entire narrative. Different from~\cite{DBLP:conf/ijcai/WangS16}, the work of~\cite{DBLP:conf/emnlp/Koncel-Kedziorski16} rewrites a given math word problem to fit a particular theme such as \textit{Star War}. In this way, students may stay more engaged with their homework assignments. The candidate are scored with the coherence in multiple factors (e.g., syntactic, semantic and thematic). \cite{DBLP:conf/ijcai/PolozovOSZGP15} generates math word problems that match the personal interest of students. The generator uses Answer Set Programming~\cite{DBLP:series/synthesis/2012Gebser}, in which programs are composed of facts and rules in a first-order logic representation, to satisfy a collection of pedagogical and narrative requirements. Its objective is to produce coherent and personalized story problems that meet pedagogical requirements.

In the branch of geometry problem generator, GeoTutor~\cite{DBLP:conf/aaai/AlvinGMM14,DBLP:journals/corr/AlvinGMM15} is designed to generate geometry proof problems for high school students. The input contains a figure and a set of geometry axioms. The output is a pair $(I,G)$, where $I$ refers to the assumptions for the figure and goals in $G$ are sets of explicit facts to be inferred. Singhal et al. also tackled the automated generation of geometry questions for high school students~\cite{DBLP:conf/csedu/SinghalHM14, DBLP:conf/csedu/SinghalHM14a}. Its input interface allows users to select geometric objects, concepts and theorems. Compared with~\cite{DBLP:conf/aaai/AlvinGMM14,DBLP:journals/corr/AlvinGMM15}, its geometric figure is generated by the algorithm rather than specified by the user. Based on the figure, the next step of generating facts and solutions is similar to that in~\cite{DBLP:conf/aaai/AlvinGMM14,DBLP:journals/corr/AlvinGMM15}. It requires pre-knowledge on axioms and theorems and results in the formation capturing the relationships between its objects.

%\input{tex/reasoning}
%%%%%%%%%%%%%%%%%%%%%%%%%%%%%%%%%%%%%%%%%%%%%%%%%%%%%%%%%%%%%%%%%

\section{Conclusions and Future Directions}
In this paper, we present a comprehensive survey to review the development of math word problem solvers in recent years. The topics discussed in this survey cover arithmetic word problems, equation set problems, geometry word problems and miscellaneous others related to math. We compared the techniques proposed for each math task, provided a rational categorization, and conducted accountable experimental analysis. Moreover, we took a close examination on the subject of feature engineering proposed for MWP solvers and summarized the diversified proposal of syntactic features. 

Overall speaking, the current status of MWP solvers still has great room for improvement. There is no doubt that the topic would continue to attract more and more research attention in the next few years, especially after the public release of large-scale datasets such as Dolphin18K and Math23K.  In the following, we present a number of possible future directions that may be of interest to the community.

Firstly, DNS~\cite{DBLP:conf/emnlp/WangLS17} was the first attempt that used deep learning models in MWP solvers so as to avoid non-trivial feature engineering. This work shed light on the feasibility of designing end-to-end models to enhance the accuracy and reduce human intervention. We observed that there have been a number of publications following this direction. For example, T-RNN is a recent work which uses Bi-LSTM and self-attention to generate quantity representation and applies recursive neural networks to infer the unknown variables in the expression tree. 

%Considering that DNS only uses a basic seq-to-seq network structure, with LSTM and GRU as the encoding and decoing units, we expect more advanced networks to be developed. 

% Moreover, as a common practice, these models can be integrated with attention mechanism~\cite{DBLP:journals/corr/BahdanauCB14} for performance advancement.

Secondly, aligning visual understanding with text mention is an emerging direction that is particularly important to solve geometry word problems. However, this challenging problem has only been evaluated in self-collected and small-scale datasets, similar to those early efforts on evaluating the accuracy of solving algebra word problem. There is a  chance that these proposed aligning methods fail to work well in a large and diversified dataset. Hence, it calls for a new round of evaluation for generality and robustness with a better benchmark dataset for geometry problems.

Thirdly, interpretability plays a key role in measuring the usability of MWP solvers in the application of online tutoring, but may pose new challenges for the deep learning based solvers~\cite{DBLP:conf/emnlp/WangTGZBL16}. For instance, AlphaGo~\cite{44806} and AlphaZero~\cite{Silver2017Mastering} have achieved astonishing superiority over human players, but their near-optimal actions could be difficult for human to interpret. Similarly, for MWP solvers, domain knowledge and reasoning capability are useful and they are friendly for interpretation. It may be interesting to combine the merits of DL models, domain knowledge and reasoning capability to develop more powerful MWP solvers.

%It would be interesting to combine the merits of DL models leverage domain knowledge and reasoning capability in the model design.  

%Furthermore, domain knowledge and reasoning capability are important. It would be interesting to 

Last but not the least, solving math word problems in English plays a dominating role in the literature. We only observed a very rare number of math solvers proposed to cope with other languages. This research topic may grow into a direction with significant impact. To our knowledge, many companies in China have harvested an enormous number of word problems in K12 education. As reported in 2015\footnote{http://www.marketing-interactive.com/baidus-zuoyebang-attracts-outside-investors/}, Zuoyebang, a spin off from Baidu, has collected $950$ million questions and solutions in its database. When coupled with deep learning models, this is an area with immense imagination and exciting achievements can be expected.
%\item policy network based reasoning; AAAI best paper

\bibliographystyle{IEEEtran}
\bibliography{ref}

\begin{IEEEbiography}[{\includegraphics[width=0.9in,height=1in,clip,keepaspectratio]{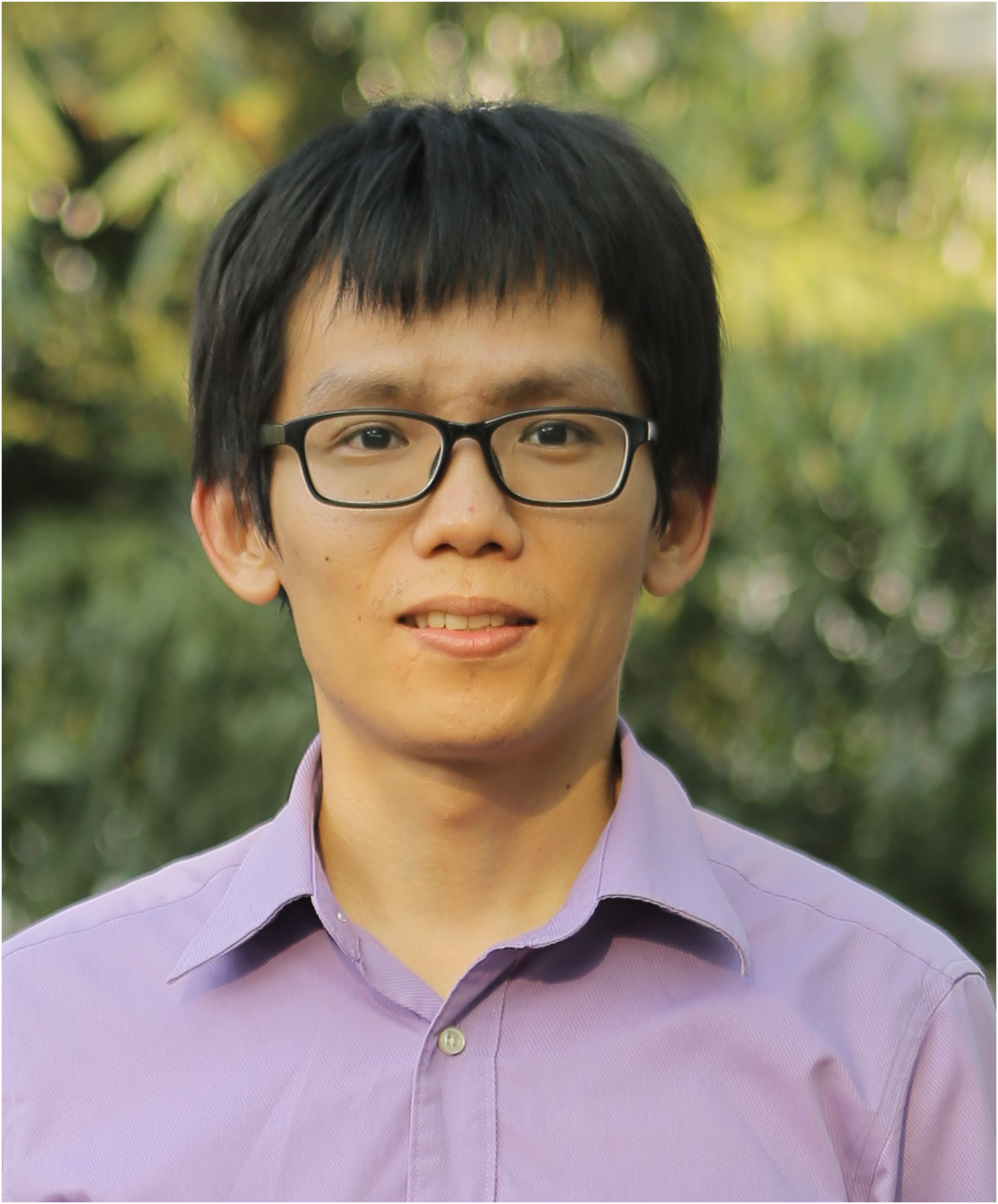}}]{Dongxiang Zhang} is a research in Zhejiang University, China. He received the B.Sc. degree from Fudan University, China in 2006 and the PhD degree from National University of Singapore in 2012. He worked as a research fellow at the NeXT research center in Singapore from 2012 to 2014,
and he was promoted to senior research fellow in 2015. His current research interests include smart education and natural language processing.
\end{IEEEbiography}

\begin{IEEEbiography}[{\includegraphics[width=0.9in,height=1in,clip,keepaspectratio]{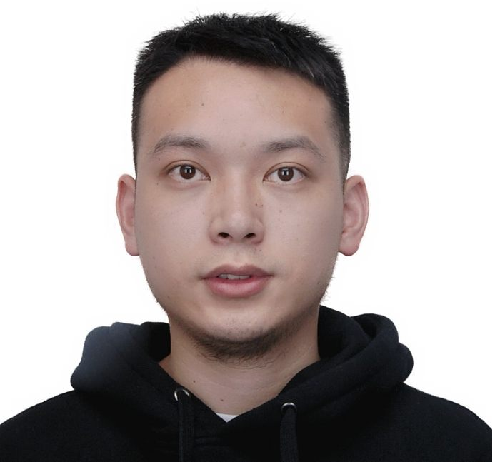}}]{Lei Wang} is currently a graduate student in the University of Electronic Science and Technology of China. He has been a research intern in Tencent AI Lab, Singapore Management University,  and Afanti Research.  His research interests mainly focus on nature language processing, reinforcement learning and machine learning. 
\end{IEEEbiography}

\begin{IEEEbiography}[{\includegraphics[width=0.9in,height=1in,clip,keepaspectratio]{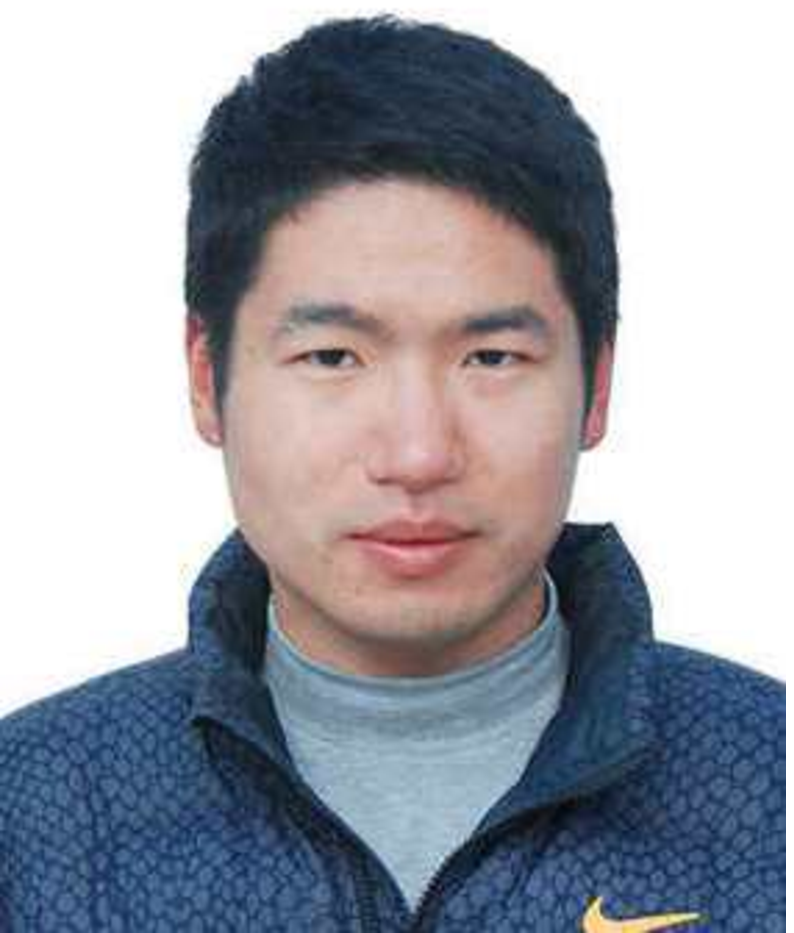}}]{Luming Zhang} (M'14) is a researcher at Zhejiang University, China. His research interests include image enhancement and pattern recognition. 
He has authored/co-authored more than 70 scientific articles at top venues including IEEE T-IP, T-MM, T-CYB, CVPR ACM MM,
 IJCAI and AAAI. He was the program committee member/organizer of many international conferences, such as MMM, PCM, ACM Multimedia. He is/was 
the associate editor of many international journals like Neurocomputing and KSII Transactions on Internet and Information Systems.

\end{IEEEbiography}

\begin{IEEEbiography}[{\includegraphics[width=0.9in,height=1in,clip,keepaspectratio]{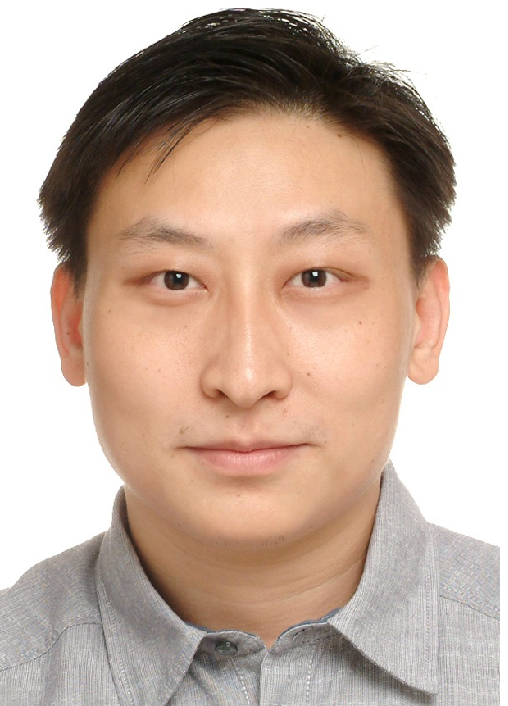}}]{Bing Tian Dai} is currently an assistant professor at the School of Information Systems, Singapore Management University. He is also the director of MITB (Artificial Intelligence) Programme. Bing Tian received his PhD from National University of Singapore in 2011 with research works on applying machine learning techniques to solve database problems for data mining applications. Prior to his faculty position, he was a research scientist at Living Analytics Research Centre, working on graph mining and machine learning problems on social media and social networks.
\end{IEEEbiography}

\begin{IEEEbiography}[{\includegraphics[width=0.9in,height=1in,clip,keepaspectratio]{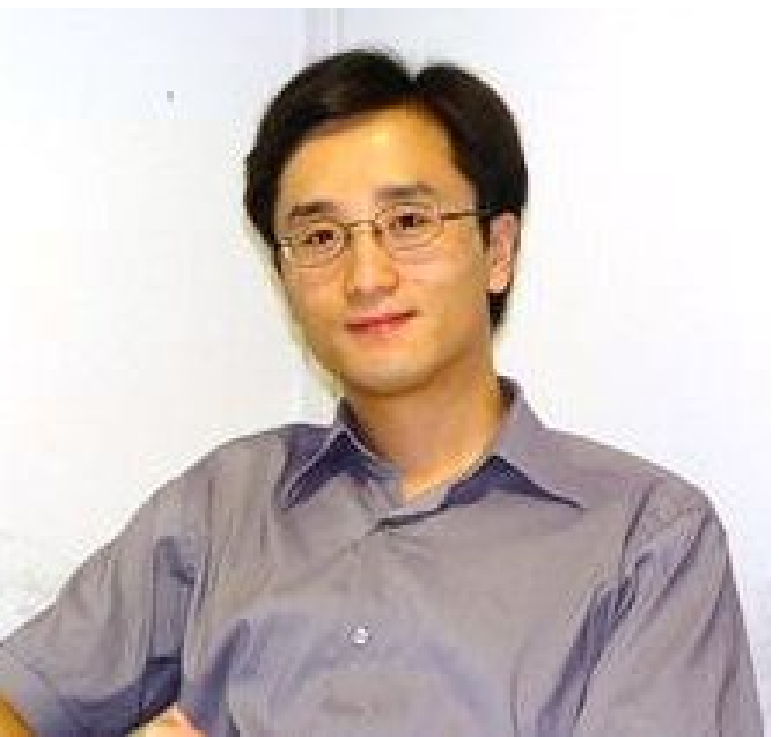}}]{Heng Tao Shen} is currently a Professor of National ``Thousand Talents Plan'', the Dean of School of Computer Science and Engineering, and the Director of Center for Future Media at the University of Electronic Science and Technology of China. He is also an Honorary Professor at the University of Queensland. He obtained his BSc with 1st class Honours and PhD from Department of Computer Science, National University of Singapore in 2000 and 2004 respectively. He then joined the University of Queensland as a Lecturer, Senior Lecturer, Reader, and became a Professor in late 2011. His research interests mainly include Multimedia Search, Computer Vision, Artificial Intelligence, and Big Data Management. He has published 200+ peer-reviewed papers, most of which appeared in top ranked publication venues, such as ACM Multimedia, CVPR, ICCV, AAAI, IJCAI, SIGMOD, VLDB, ICDE, TOIS, TIP, TPAMI, TKDE, VLDB Journal, etc. He has received 6 Best Paper Awards from international conferences, including the Best Paper Award from ACM Multimedia 2017 and Best Paper Award - Honorable Mention from ACM SIGIR 2017. He has served as a PC Co-Chair for ACM Multimedia 2015 and currently is an Associate Editor of IEEE Transactions on Knowledge and Data Engineering. 
\end{IEEEbiography}

\end{document}